\documentclass[11pt]{article}
\pdfoutput=1
\usepackage{longtable}
\usepackage{wrapfig}
\usepackage[utf8]{inputenc} 
\usepackage[T1]{fontenc}    
\usepackage[colorlinks,
            linkcolor=red,
            anchorcolor=blue,
            citecolor=blue
            ]{hyperref}
\usepackage{url}            
\usepackage{booktabs}       
\usepackage{amsfonts}       
\usepackage{nicefrac}       
\usepackage{xcolor,colortbl}         
\usepackage{wrapfig}
\usepackage{caption}
\usepackage{enumitem}
\usepackage{tabularx}
\usepackage{mylatexstyle}
\usepackage[compact]{titlesec}
\usepackage{multirow}
\usepackage{setspace}
\usepackage{fullpage}

\usepackage[protrusion=true,expansion=true]{microtype}
\usepackage{pbox}
\usepackage{setspace}
\usepackage{tabularx}
\usepackage{float}
\usepackage{wrapfig,lipsum}
\usepackage{enumitem}
\usepackage{graphicx}
\usepackage{subfigure}
\usepackage{pgfplots}
\usetikzlibrary{arrows,shapes,snakes,automata,backgrounds,petri}
\usepackage{booktabs} 

\allowdisplaybreaks
\usepackage{colortbl}

\usepackage{enumitem}
\usepackage{colortbl}
\usepackage{graphicx} 
\usepackage{algorithm}
\usepackage{algpseudocode}  
\usepackage{tikz}
\usetikzlibrary{positioning, arrows.meta, calc, fit, backgrounds, shadows.blur, shapes.geometric, shadows}
\ifdefined\final
\usepackage[disable]{todonotes}
\else
\usepackage[textsize=tiny]{todonotes}
\fi
\definecolor{aBlue}    {HTML}{007AFF}
\definecolor{aIndigo}  {HTML}{5856D6}
\definecolor{aPurple}  {HTML}{AF52DE}
\definecolor{aPink}    {HTML}{FF2D55}
\definecolor{aRed}     {HTML}{FF3B30}
\definecolor{aOrange}  {HTML}{FF9500}
\definecolor{aYellow}  {HTML}{FFCC00}
\definecolor{aGreen}   {HTML}{34C759}
\definecolor{aTeal}    {HTML}{30B0C7}
\definecolor{aGray}    {HTML}{8E8E93}
\definecolor{aGrayLn} {HTML}{C7C7CC}
\definecolor{aBg}      {HTML}{F2F2F7}
\definecolor{aTxt}     {HTML}{1C1C1E}
\definecolor{aSub}     {HTML}{6E6E73}
\newcommand{\SG}{\mathrm{SG}}
\newcommand{\E}{\mathbb{E}}

\newcommand{\KL}{D_{\mathrm{KL}}}
\newcommand{\piT}{\pi_{\mathrm{teacher}}}
\newcommand{\piS}{\pi_{\theta}}
\newcommand{\piO}{\pi_{\mathrm{old}}}
\newcommand{\piR}{\pi_{\mathrm{ref}}}
\newcommand{\Adist}{A^{\mathrm{dist}}}

\makeatletter
\@ifundefined{proposition}{\newtheorem{proposition}{Proposition}}{}
\makeatother

\title{\huge Self-Distilled Policy Gradient}

\author
{
    Yifeng Liu\thanks{Equal contribution} \thanks{Department of Computer Science, University of California, Los Angeles, CA, USA; email: liuyifeng@cs.ucla.edu} 
	~~~
    Shiyuan Zhang\footnotemark[1] \thanks{Department of Computer Science, University of California, Los Angeles, CA, USA; email: zsy25ucla@ucla.edu} 
	~~~
    Yifan Zhang\footnotemark[1] \thanks{Princeton AI Laboratory, Princeton University, Princeton, NJ, USA; email: yifzhang@princeton.edu} 
	~~~
	Quanquan Gu\footnotemark[2] \thanks{Corresponding Author, Department of Computer Science, University of California, Los Angeles, CA, USA; email: qgu@cs.ucla.edu}
}

\date{}

\begin{document}

\maketitle

\begin{abstract} On-policy self-distillation, where a language model conditions on privileged context to supervise its own generations, is a promising source of dense supervision for sparse-reward reinforcement learning. Actually, it can be instantiated as an auxiliary full-vocabulary student-to-teacher reverse Kullback-Leibler divergence loss. We therefore propose SDPG, a self-distilled policy-gradient framework that combines group-relative verifier advantages with normalized standard deviation, exact full-vocabulary on-policy self-distillation, as well as reference-policy KL regularization. Empirically, SDPG improves stability and performance over RLVR and self-distillation baselines. The code is available at \url{https://github.com/lauyikfung/SDPG}.

\end{abstract}


\begin{figure}[h]
\centering
\resizebox{0.95\textwidth}{!}{%
\begin{tikzpicture}[
    font=\sffamily,
    >=Stealth,
    line cap=round, line join=round,
    every node/.style={text=aTxt, outer sep=0pt},
    card/.style 2 args={
        rectangle, rounded corners=5pt,
        draw=#1, line width=0.7pt,
        fill=#1!8, text=aTxt,
        align=center, inner sep=5pt,
        font=\sffamily\small,
        minimum width=#2, minimum height=10mm,
    },
    cardLg/.style 2 args={
        rectangle, rounded corners=8pt,
        draw=#1, line width=0.9pt,
        fill=#1!8, text=aTxt,
        align=center, inner sep=7pt,
        font=\sffamily\small,
        minimum width=#2,
    },
    pill/.style 2 args={
        rectangle, rounded corners=9pt,
        draw=#1, line width=0.6pt,
        fill=#1!12, text=aTxt,
        align=center, inner sep=4pt,
        font=\sffamily\footnotesize,
        minimum width=#2, minimum height=7mm,
    },
    chip/.style 2 args={
        rectangle, rounded corners=10pt,
        draw=#1, line width=0.5pt,
        fill=#1!10, text=aTxt,
        align=center, inner sep=3pt,
        font=\sffamily\scriptsize,
        minimum width=#2, minimum height=6mm,
    },
    flow/.style={->, line width=0.85pt, draw=aGrayLn,
                 shorten >=1.5pt, shorten <=1.5pt},
    flowC/.style 2 args={->, line width=#1, draw=#2,
                        shorten >=1.5pt, shorten <=1.5pt},
    note/.style={font=\sffamily\scriptsize, text=aSub, inner sep=2pt,
                 fill=white, fill opacity=0.92, text opacity=1, rounded corners=2pt},
]

\node[card={aBlue}{22mm}]   (xprompt)  at (-3.6, 4.0) {Prompt $x$};
\node[card={aPurple}{30mm}] (ccontext) at ( 3.6, 4.0) {Privileged context $c$};

\node[cardLg={aIndigo}{40mm}, minimum height=15mm]
      (model) at (0, 2.0)
      {{\large $\bm{\pi_\theta}$}\\[1pt]\textcolor{aSub}{\scriptsize shared model}};

\node[card={aBlue}{52mm}]    (pt) at (-3.8, -0.4)
      {Student~~$p_t \;=\; \pi_\theta(\,\cdot\mid x,\,y_{<t}\,)$};
\node[card={aPurple}{52mm}]  (qt) at ( 3.8, -0.4)
      {Teacher~~$q_t \;=\; \pi_\theta(\,\cdot\mid c,\,x,\,y_{<t}\,)$};

\node[card={aTeal}{40mm}]    (yroll) at (-5.6, -2.6)
      {Rollouts $\{y^{(i)}\}_{i=1}^{G}\sim p_t$};
\node[card={aGreen}{50mm}]   (verif) at (-5.6, -4.4)
      {Verifier~~$R(x, y^{(i)})$\\
       $A^{(i)}_{\text{out}} \;=\; \dfrac{R^{(i)}-\mu_G}{\sigma_G+\epsilon_{\mathrm{std}}}$};

\node[cardLg={aOrange}{56mm}, minimum height=15mm] (opd) at (3.8, -2.9)
      {\textbf{Full-Vocab OPD KL}\\[2pt]
       $\ell^{\text{OPD}}_t \;=\; D_{\text{KL}}\!\bigl(p_t\,\big\|\,\mathrm{SG}[\,q_t\,]\bigr)$};

\node[pill={aPink}{42mm}] (gate) at (-0.2, -4.4)
      {gate~~$m_i \;=\; \mathbf{1}\!\bigl[A^{(i)}_{\text{out}} > 0\bigr]$};

\node[card={aGreen}{40mm},  minimum height=12mm] (lout) at (-5.4, -6.6)
      {$\mathcal L_{\text{out}}$\\[1pt]
       \textcolor{aSub}{\scriptsize on-policy policy gradient}};

\node[card={aOrange}{40mm}, minimum height=12mm] (lopd) at (-0.2, -6.6)
      {$\beta(k)\,\mathcal L^{+}_{\text{OPD}}$\\[1pt]
       \textcolor{aSub}{\scriptsize gated~+~scheduled}};

\node[card={aRed}{40mm},    minimum height=12mm] (lkl)  at ( 5.0, -6.6)
      {$\alpha\,\mathcal L_{\mathcal K}(\pi_\theta,\pi_{\text{ref}})$\\[1pt]
       \textcolor{aSub}{\scriptsize reference KL regularization (UFKL/URKL)}};

\node[chip={aRed}{16mm}] (piref) at (8.5, -4.4) {$\pi_{\text{ref}}$\\[-1pt]\textcolor{aSub}{fixed}};

\node[cardLg={aIndigo}{135mm}, minimum height=14mm] (lsdpg) at (-0.2, -8.5)
      {{\large $\bm{\mathcal L_{\text{SDPG}}}\;=\;\mathcal L_{\text{out}}\;+\;\beta(k)\,\mathcal L^{+}_{\text{OPD}}\;+\;\alpha\,\mathcal L_{\mathcal K}\bigl(\pi_\theta,\pi_{\text{ref}}\bigr)$}};

\draw[flow] (xprompt.south)  -- ++(0,-0.2) -| ([xshift=-9mm]model.north);
\draw[flow] (ccontext.south) -- ++(0,-0.2) -| ([xshift= 9mm]model.north);

\draw[flowC={0.95pt}{aBlue!70}]
      ([xshift=-12mm]model.south) -- ++(0,-0.5) -| (pt.north);
\node[note] at (-2.6, 0.85) {without $c$};

\draw[flowC={0.95pt}{aPurple!70}]
      ([xshift= 12mm]model.south) -- ++(0,-0.5) -| (qt.north);
\node[note] at ( 2.6, 0.85) {with $c$};

\draw[flow] ([xshift=-18mm]pt.south) -- (yroll.north);
\draw[flow] (yroll.south) -- (verif.north);

\draw[flow] (pt.east) -- ++(0.4,0) |- (opd.west);
\draw[flow] (qt.south) -- (opd.north);

\draw[flow] (verif.east) -- (gate.west);

\draw[flow] (verif.south) -- ++(0,-0.45) -| (lout.north);

\draw[flow] (opd.south) -- ++(0,-1.5) -| ([xshift=15mm]lopd.north);

\draw[flow] (gate.south) -- (lopd.north) node[note, midway, left=2pt] {modulates};

\draw[flow] (piref.south) -- ++(0,-0.6) -| (lkl.north);

\draw[flowC={1.0pt}{aGreen!85}]  (lout.south) -- (lout.south  |- lsdpg.north);
\draw[flowC={1.0pt}{aOrange!85}] (lopd.south) -- (lopd.south  |- lsdpg.north);
\draw[flowC={1.0pt}{aRed!85}]    (lkl.south)  -- (lkl.south   |- lsdpg.north);

\node[font=\sffamily\scriptsize, text=aSub, anchor=east] at (-7.8, 4.0) {\textsc{inputs}};
\node[font=\sffamily\scriptsize, text=aSub, anchor=east] at (-7.8, 2.0) {\textsc{policy}};
\node[font=\sffamily\scriptsize, text=aSub, anchor=east] at (-7.8,-0.4) {\textsc{distributions}};
\node[font=\sffamily\scriptsize, text=aSub, anchor=east] at (-7.8,-3.5) {\textsc{signals}};
\node[font=\sffamily\scriptsize, text=aSub, anchor=east] at (-7.8,-6.6) {\textsc{losses}};
\node[font=\sffamily\scriptsize, text=aSub, anchor=east] at (-7.8,-8.5) {\textsc{objective}};

\begin{scope}[on background layer]
    \fill[aBg, rounded corners=10pt]
        (-7.5, 3.4) rectangle (7.5, 4.65);
    \fill[aBg, rounded corners=10pt]
        (-7.5,-7.35) rectangle (7.5,-5.85);
\end{scope}

\end{tikzpicture}%
}
\caption{Overview of the Self-Distilled Policy Gradient (SDPG) objective, combining rollout-based outcome optimization, gated On-Policy Distillation (OPD) from privileged context, and a KL regularization to a fixed reference policy. Note that OPD is also a form of policy gradient.}
\label{fig:sdpg-overview}
\end{figure}
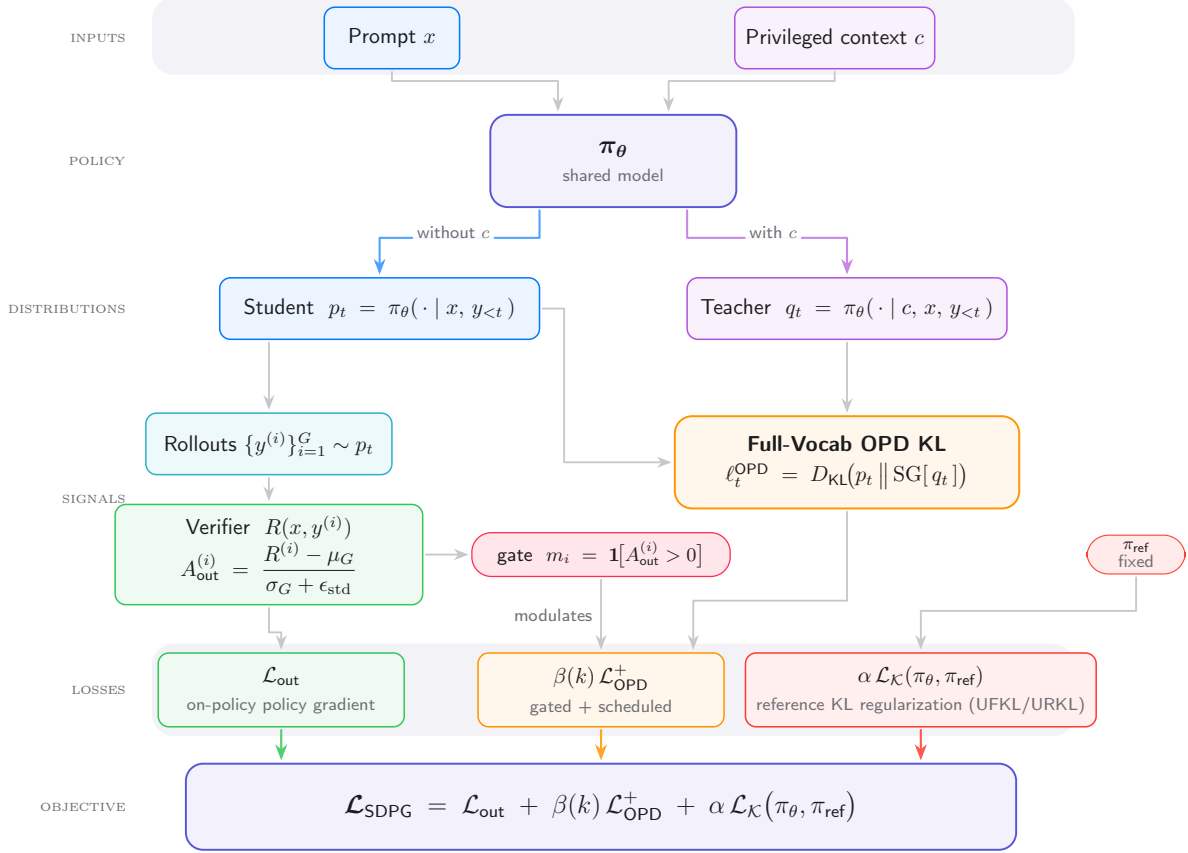

\section{Introduction}

With the development of Reinforcement Learning with Verifiable Rewards (RLVR), Large Language Models (LLMs) have demonstrated remarkable capabilities in complex reasoning tasks such as mathematics and code generation. Algorithms in this family, such as Group Relative Policy Optimization (GRPO)~\citep{shao2024deepseekmath}, optimize against rule-based outcome rewards and have become the standard recipe for post-training reasoning models, eliminating the cost and bias of human preference annotation.

Despite its success, RLVR encounters several limitations, including sparse sequence-level reward across tokens and instability under negative advantages during the early stages of training. Although recent works such as Dr.GRPO~\citep{liu2025understanding}, DAPO~\citep{yu2025dapo}, and GSPO~\citep{zheng2025group} address the latter through asymmetric dual-clip thresholds and sequence-level advantage, the sparsity issue remains unresolved.

Recently, on-policy distillation (OPD) approaches have been proposed to yield dense token-level signals~\citep{agarwal2024policy,lu2025onpolicy,fu2026revisiting}. Such methods maintain two models: a student model to be optimized that rolls out trajectories, and a teacher model that produces token-level guidance via Kullback–Leibler divergence
(KL) regularization or related objectives~\citep{gu2024minillm,xu2025speculative,yang2025qwen3}. However, traditional distillation approaches use a much larger and stronger teacher, which imposes a considerable memory burden when optimizing student models. Moreover, heterogeneous teacher signals may hurt the smoothness of the training process.

A recent line of work addresses such limitations through on-policy self-distillation. In these methods, the teacher model is exactly the same model as the student model, but with additional knowledge such as demonstrations, direct answers, and reasoning paths~\citep{hubotter2026reinforcement,shenfeld2026self,penaloza2026privileged}. This converts sparse and inconsistent outcome rewards into dense, per-token, and homogeneous supervision. In detail, OPCD~\citep{ye2026policy} involves in-context knowledge in the teacher model and internalizes it into the student model through KL divergence; OPSD~\citep{zhao2026self} applies a full, vocabulary-wise KL divergence for better reasoning performance; and TRRD~\citep{zhang2026reinforcement} incorporates trust regions in distillation.

However, the phrase ``self-distillation'' can obscure a useful policy-gradient interpretation. For a fixed rollout prefix $(x,y_{<t})$, let
\[
    q_t(a)=\piS(a\mid c,x,y_{<t})
    \quad\text{and}\quad
    p_t(a)=\piS(a\mid x,y_{<t}),
\]
where $q_t$ is the privileged distribution induced by the same model under context $c$, and $p_t$ is the deployable distribution without privileged context. We use the full-vocabulary reverse KL $\KL(p_t\|\SG[q_t])$ between these two distributions. Conditioned on the sampled prefix and with the privileged branch detached, its student-side gradient is locally identical to a detached-sampling policy-gradient update whose token advantage is a centered log teacher/student ratio. This is a gradient identity, not a replacement of the full-vocabulary KL objective. It also suggests a failure mode of pure self-distillation: without a verifier, a privileged but imperfect distribution can reinforce locally plausible tokens on globally wrong trajectories. A natural remedy is to combine the advantages of both self-distillation and RLVR methods by keeping the GRPO reward objective with self-distillation signals. RLSD~\citep{yang2026self} incorporates self-distillation in the clipped importance ratio of GRPO loss function to moderate the teacher signal. Nevertheless,~\citep{xiao2026mimo} still notices that strong teacher signals may hurt the potential of the student models. 

In order to address the above challenge, we propose \textbf{SDPG} (Self-Distilled Policy Gradient), which integrates exact full-vocabulary privileged OPD into KL-regularized policy optimization. The resulting objective has two complementary sources of supervision: a sparse binary outcome signal from the verifier and a dense full-vocabulary distillation signal from the context-conditioned teacher. Based on the RPG framework~\citep{zhang2026design}, we focus on unnormalized reference-policy KL regularizations in the main text, with normalized variants and all anchor derivations deferred to the appendix. To control noise in the privileged teacher signal, we apply positive-advantage gating and a warmup-decay schedule to the distillation coefficient. Our contributions are:
\begin{enumerate}[leftmargin=*,itemsep=1pt,topsep=1pt]
    \item We derive an exact local policy-gradient form of reverse-KL full-vocabulary OPD: for privileged teacher $q_t$ and deployable student $p_t$, the student-side gradient of $D_{\mathrm{KL}}(p_t\|\SG[q_t])$ equals a detached-sampling update with centered log-ratio advantage $\SG[\bar D_t-\log(\bar p_t(a)/\bar q_t(a))]$, where $\bar D_t=\KL(\bar p_t\|\bar q_t)$.
    \item We propose a KL-regularized policy optimization objective that combines binary outcome rewards with exact full-vocabulary privileged distillation, and instantiate it with rollout-policy-sampled unnormalized forward and reverse KL regularizations to a fixed reference policy.
    \item We develop two stabilizers, positive-advantage gating and a warmup-decay schedule for the distillation coefficient, and show that SDPG improves over GRPO and self-distillation baselines.
\end{enumerate}

\section{Background}

Before introducing the SDPG framework, we briefly review GRPO, one of the standard paradigms in RLVR and introduce on-policy self-distillation. After that, we formally define the divergence measures, including the unnormalized KL divergence, which serves as the theoretical foundation for our main KL regularizations.

\subsection{Group Relative Policy Optimization}

In reasoning tasks, such as mathematics or coding, models are typically trained using verifiable outcome-based rewards. Given a prompt $x$, the model generates a sequence of tokens $y=\{y_1,y_2,\dots,y_{|y|}\}$. A rule-based verifier assigns a scalar reward $R(x,y)$, e.g., $1$ for correct and $0$ for incorrect.

Group Relative Policy Optimization (GRPO)~\citep{shao2024deepseekmath} is the current standard for RLVR. For each prompt $x$, GRPO samples a group of $G$ outputs $\{y^{(1)},\dots,y^{(G)}\}$ from a frozen rollout policy $\pi_{\mathrm{old}}$, usually the current policy snapshot at the beginning of the update. Instead of training a separate value network, GRPO computes a sequence-level advantage $A^{(i)}$ by normalizing the rewards within the group:
\begin{align}
    A^{(i)} = \frac{R(x,y^{(i)})-\mu_G}{\sigma_G+\varepsilon_{\mathrm{std}}},
    \label{eq:grpo_advantage}
\end{align}
where $\mu_G$ and $\sigma_G$ are the mean and standard deviation of the group's rewards, and $\varepsilon_{\mathrm{std}}>0$ avoids division by zero. Equivalently, implementations often set $A^{(i)}=0$ when all rewards in the group are identical. The policy is then optimized using a PPO-style~\citep{schulman2017proximal} clipped surrogate objective:
\begin{align}
\resizebox{1.0\textwidth}{!}{$
\begin{aligned}
 \mathcal{L}_{\text{GRPO}}(\theta)
 =
 -\mathbb{E}_{x,\;\{y^{(i)}\}_{i=1}^G\sim\pi_{\mathrm{old}}(\cdot\mid x)}
 \bigg[
 \frac{1}{\sum_i |y^{(i)}|}
 \sum_{i=1}^{G}
 \sum_{t=1}^{|y^{(i)}|}
 \min\Bigl(
 r_{i,t} A^{(i)},\;
 \mathrm{clip}(r_{i,t},1-\epsilon,1+\epsilon)A^{(i)}
 \Bigr)
 \bigg],
 \label{eq:grpo_loss}
\end{aligned}
$}
\end{align}
where $r_{i,t}$ is the importance ratio defined as follows:
\[
r_{i,t}=
\frac{\pi_\theta(y^{(i)}_t\mid x,y^{(i)}_{<t})}
{\pi_{\text{old}}(y^{(i)}_t\mid x,y^{(i)}_{<t})}
\]
is the importance sampling ratio. During the policy update, $\pi_{\mathrm{old}}$ is fixed and gradients are taken only through $\pi_\theta$.

While GRPO is computationally efficient, the scalar advantage $A^{(i)}$ is applied equally to every token in the sequence. This sparse, sequence-level credit assignment makes it difficult for the model to identify which specific reasoning steps were correct or flawed, leading to inefficient exploration. Furthermore, standard PPO clipping struggles with the overwhelming number of negative advantages ($A<0$) early in training, which can degrade the model's foundational language capabilities.

\subsection{On-Policy Self-Distillation}

In standard off-policy knowledge distillation, a smaller student model is trained to mimic the behavior of a more capable, distinct teacher model. The conventional objective minimizes the divergence between the teacher's and student's output distributions over trajectories generated by the teacher. While effective, this off-policy paradigm suffers from exposure bias and distribution mismatch, as the student is trained on the teacher's distribution but evaluated on its own rollouts during inference.

On-policy self-distillation reduces distribution mismatch by evaluating the distillation loss on prefixes sampled from the student's rollout distribution, usually a frozen snapshot of the current policy. It also eliminates the need for an external teacher model: a single model $\pi_\theta$ acts as the deployable student when conditioned on $x$ and as the privileged teacher when additionally conditioned on $c$, such as ground-truth reasoning traces or verified reference answers. At prefix $(x,y_{<t})$, the two distributions are
\begin{align*}
    q_t(a)=\pi_\theta(a\mid c,x,y_{<t}),
    \qquad
    p_t(a)=\pi_\theta(a\mid x,y_{<t}).
\end{align*}
Prior work usually aligns these distributions through a full-vocabulary KL divergence. In this work we use the student-to-teacher reverse KL $\KL(p_t\|\SG[q_t])$. Proposition~\ref{prop:opd_advantage} shows that, on a fixed sampled prefix and with the teacher branch detached, its student-side gradient has an equivalent local policy-gradient form.

\subsection{Unnormalized KL Divergence}
\label{sec:unnormalized_KL}

The standard Normalized Kullback-Leibler (KL) divergence between two distributions $P$ and $Q$ is defined as:
\begin{align*}
    D_{\text{KL}}(P \| Q) = \int P(x) \log \frac{P(x)}{Q(x)} dx.
\end{align*}
However, in the scenario when reference measures may not be perfectly normalized~\citep{zhang2026design} (i.e., $\int_x P(x)$ and $\int_x Q(x)$ may be not equal to 1), to yield a more elegant and symmetric gradient, we employ the Unnormalized KL (UKL) divergence~\citep{zhu1995information,minka2005divergence}. UKL introduces a mass correction term to handle scenarios where the distributions sum over sub-vocabularies or are otherwise unnormalized. It is defined as:
\begin{align}
    \text{UKL}(P \| Q) = \underbrace{\int P(x) \log \frac{P(x)}{Q(x)} dx}_{D_{\text{KL}}(P \| Q)} + \underbrace{\int \big( Q(x) - P(x) \big) dx}_{\text{Mass Correction}}.~\label{eq:ukl}
\end{align}
Actually, it is equivalent to $k_3$ estimator~\citep{Schulman_KLApprox}, which is an unbiased estimator of KL divergence but with lower variance. While we will employ the exact token-level KL over the full vocabulary for the self-distillation objective, we will use this unnormalized KL (UKL) framework for the policy regularization against the reference model $\piR$.

\section{Self-Distilled Policy Gradient}
\label{sec:method}

SDPG has three components. First, it uses on-policy objective as used in RLVR, with rewards supplied by a binary verifier. Second, it adds exact full-vocabulary OPD on prefixes sampled from the unprivileged rollout policy, so the privileged signal is dense but remains tied to the student's own state distribution. Third, it anchors the updated policy to a fixed reference policy through a rollout-policy-sampled KL surrogate. Algorithm~\ref{alg:SDPG-self} summarizes the training loop.

\begin{algorithm}[htb]
\small
\caption{SDPG: Self-Distilled Policy Gradient with Full-Vocabulary OPD}
\label{alg:SDPG-self}
\begin{algorithmic}
\Require Training data $\mathcal{D}=\{(x,c)\}$, where $x$ is the input and $c$ is privileged in-context knowledge; language model $\pi_\theta$; fixed reference policy $\piR$; total training steps $T$; $\epsilon_\mathrm{std}>0$.
\For{each training step $k=1,\ldots,T$}
    \State Sample a batch of prompts and privileged contexts $\{(x_j,c_j)\}_{j=1}^{B}\sim\mathcal D$
    \For{each prompt $x_j$}
        \State \textcolor{gray}{\textit{// Rollout from the frozen unprivileged behavior policy}}
        \State Sample a group of $G$ responses $\{y_j^{(i)}\}_{i=1}^{G}\sim \piS(\cdot\mid x_j)$
        \State \textcolor{gray}{\textit{// Compute outcome rewards and group-relative advantages}}
        \State Obtain binary verifier rewards $R_j^{(i)}=R(x_j,y_j^{(i)})$
        \State Compute $A^{(i)}_{\mathrm{out}}=\frac{R_j^{(i)}-\mu_j}{\sigma_j+\epsilon_\mathrm{std}}$, where $\mu_j$ and $\sigma_j$ are the mean and standard deviation of $\{R_j^{(i)}\}_{i=1}^G$
        \State Set $m_j^{(i)}=\mathbf 1[A^{(i)}_{\mathrm{out}}>0]$
        \For{each response $y_j^{(i)}$ and token position $t$}
            \State Define the prefix state $s_{j,i,t}=(x_j,y_{j,<t}^{(i)})$
            \State \textcolor{gray}{\textit{// Student, privileged-teacher, behavior, and reference distributions}}
            \State Compute $p_{j,i,t}=\pi_\theta(\cdot\mid s_{j,i,t})$, $q_{j,i,t}=\SG[\pi_\theta(\cdot\mid c_j,s_{j,i,t})]$ and $\piR(\cdot\mid s_{j,i,t})$
            \State \textcolor{gray}{\textit{// Exact full-vocabulary OPD loss on this sampled prefix}}
            \State $\mathcal{L}^{\mathrm{OPD}}_{j,i,t}=\sum_{a\in\mathcal V}p_{j,i,t}(a)\log\frac{p_{j,i,t}(a)}{q_{j,i,t}(a)}$
        \EndFor
    \EndFor
    \State Update $\theta$ by minimizing $\mathcal{L}_\mathrm{SDPG}(\theta)$ in Eq.~\eqref{eq:SDPG_loss}
\EndFor
\State \Return $\pi_\theta$
\end{algorithmic}
\end{algorithm}

\subsection{KL-Regularized Policy Optimization with Outcome and OPD}

SDPG minimizes a KL-regularized policy optimization objective with two sources of supervision:
\begin{align}
\mathcal L_{\mathrm{SDPG}}(\theta)
=
\mathcal L_{\mathrm{out}}(\theta)
+
\beta(k)\mathcal L_{\mathrm{OPD}}^{+}(\theta)
+
\alpha\mathcal L_{\mathcal K}(\pi_\theta,\piR),
\label{eq:SDPG_loss}
\end{align}
where $\mathcal L_{\mathrm{out}}$ is the reward-based policy-gradient loss, $\mathcal L_{\mathrm{OPD}}^{+}$ is the gated full-vocabulary OPD loss, $\beta(k)$ is the distillation coefficient at training step $k$, and $\mathcal L_{\mathcal K}$ is a KL regularization against the fixed reference policy. When $\beta=0$, SDPG reduces to the corresponding RPG-style objective. However, the forms of $\mathcal{K}$ for the on-policy loss are quite different from~\citet{zhang2026design}, which is under a one-step off-policy setting. When $\alpha=0$, SDPG becomes outcome-reward policy optimization with full-vocabulary OPD but without a reference-policy anchor.

\subsection{On-policy Reward-based Loss}
\label{sec:strict_onpolicy_sdpg}

We first derive the reward-based loss $\mathcal{L}_\mathrm{out}$ in Eq.~\eqref{eq:SDPG_loss}. For full on-policy specialization, the rollout policy is exactly the current policy $\pi_\theta$. Therefore, $y_i\sim \SG[\piS(\cdot\mid x)]$, where $\SG$ is the stop-gradient operator.
For the objective 
\begin{align*}
    J_\mathrm{out}(\theta)=\EE_{x,y\sim\piS(\cdot|x)}[R(y)],
\end{align*}
to generate gradient signals from reward, we should change to REINFORCE-style surrogate loss~\cite{williams1992simple}:
\begin{align*}
    \mathcal{L}_\mathrm{out}(x_i,y_i,\theta)=-\SG[R(y_i)]\log\piS(x_i)
\end{align*}
so that $\EE_{x\sim\piS}[\nabla_\theta\mathcal{L}_\mathrm{out}(x,y,\theta)]=-\nabla_\theta J_\mathrm{out}(\theta)$. Therefore, no PPO-style importance-ratio clipping is needed. Moreover, similar as GRPO~\citep{shao2024deepseekmath}, to mitigate the variance and neutralize baseline bias, we use the same group-relative advantage $A_\mathrm{out}(y)$ as in Eq.~\eqref{eq:grpo_advantage} instead of the original reward $R(y)$. Therefore, the verifier-grounded objective can be written directly as
\begin{align*}
\mathcal L_{\mathrm{out}}(\theta)
&=
-
\E_{(x,c)\sim\mathcal D,\;\{y_i\}_{i=1}^{G}\sim\piS(\cdot\mid x)}
\bigg[
\frac{1}{\sum_{i=1}^{G}|y_i|}
\sum_{i=1}^{G}
\sum_{t=1}^{|y_i|}
\SG[A_\mathrm{out}(y_i)]\,
\log \pi_\theta(y_{i,t}\mid x,y_{i,<t})
\bigg].
\end{align*}

\subsection{Full-Vocabulary Distillation on Sampled Prefixes}
\label{sec:full_vocab_distillation}

Instead of using a sampled-token approximation to the privileged teacher signal, SDPG uses the exact full-vocabulary student-to-teacher KL on each sampled prefix similar to~\citet{zhao2026self}. For a sampled response $y_i$ and token position $t$, define the prefix $s_{i,t}=(x,y_{i,<t})$ and the two next-token distributions
\begin{align*}
    p_{i,t}(a)&=\pi_\theta(a\mid x,y_{i,<t}),\\
    q_{i,t}(a)&=\pi_\theta(a\mid c,x,y_{i,<t}).
\end{align*}
With the sampled prefix and teacher branch detached, the per-token OPD loss is
\begin{align}
    \ell^{\mathrm{OPD}}_{i,t}(\theta)
    =
    \KL(p_{i,t}\|\SG[q_{i,t}])
    =
    \sum_{a\in\mathcal V}
    p_{i,t}(a)
    \log
    \frac{p_{i,t}(a)}{\SG[q_{i,t}(a)]}.
    \label{eq:opd_token_loss}
\end{align}

\begin{proposition}[Fixed-prefix reverse-KL OPD gradient as a policy gradient]
\label{prop:opd_advantage}
Fix a rollout prefix $s_t=(x,y_{<t})$ and write
$p_t(a)=\pi_\theta(a\mid x,y_{<t})$ and
$q_t(a)=\pi_\theta(a\mid c,x,y_{<t})$. Let
$\bar p_t=\SG[p_t]$ and $\bar q_t=\SG[q_t]$, and assume
$\bar q_t(a)>0$ whenever $\bar p_t(a)>0$. With the teacher branch detached, the reverse-KL full-vocabulary OPD loss
\begin{align*}
    \mathcal L_{\mathrm{OPD},t}(\theta)
    =
    \KL(p_t\|\bar q_t)
\end{align*}
has the same student-side gradient, at the current iterate, as the detached-sampling policy-gradient surrogate
\begin{align}
    \widetilde{\mathcal L}^{\mathrm{PG}}_{\mathrm{OPD},t}(\theta)
    =
    -
    \E_{a\sim \bar p_t}
    \left[
        \Adist_t(a)\log p_t(a)
    \right],
    \qquad
    \Adist_t(a)
    =
    \SG\left[
        \bar D_t-
        \log\frac{\bar p_t(a)}{\bar q_t(a)}
    \right]\label{eq:opd_advantage_pg},
\end{align}
where $\bar D_t=
    \KL(\bar p_t\|\bar q_t)$. Moreover, $\Adist_t$ is centered under the detached student distribution:
$\E_{a\sim \bar p_t}[\Adist_t(a)]=0$.
\end{proposition}

The proof is given in Appendix~\ref{sec:opd_equivalent_policy_gradient}. Proposition~\ref{prop:opd_advantage} is a gradient identity, not an implementation change: SDPG minimizes the explicit full-vocabulary KL in Eq.~\eqref{eq:opd_token_loss}, because this leads to more accurate estimation of the gradient. The total distillation loss over sampled sequences is
\begin{align}
    \mathcal L_{\mathrm{OPD}}(\theta)
    =
    \E_{(x,c)\sim\mathcal D,\;\{y_i\}_{i=1}^{G}\sim\piS(\cdot\mid x)}
    \bigg[
    \frac{1}{\sum_{i=1}^{G}|y_i|}
    \sum_{i=1}^{G}
    \sum_{t=1}^{|y_i|}
    \ell^{\mathrm{OPD}}_{i,t}(\theta)
    \bigg].
    \label{eq:opd_loss}
\end{align}

\subsection{On-policy Unnormalized KL for SDPG}
Now we focus on the KL regularization term $\mathcal{L}_\mathcal{K}$ in Eq.~\eqref{eq:SDPG_loss}. We postpone the derivations for general forward and reverse KL regularization to Appendix~\ref{sec:normalized_KL}. From the derivation there, it is worth noting that $\piR=\piS$ is not sufficient to minimize the surrogate loss for forward and reserve KL, which is due to the inherent biases within (normalized) forward and backward KL regularization. To tackle this mismatching issue, we apply the unnormalized KL term as introduced in Section~\ref{sec:unnormalized_KL}. For simplicity, we denote $J_\mathrm{R\&D}=J_\mathrm{out}+\beta(k) J_\mathrm{OPD}$ and $\mathcal{L}_\mathrm{R\&D}=\mathcal{L}_\mathrm{out}+\beta(k)\mathcal{L}_\mathrm{OPD}$ for training step $k$ as the objective and loss functions for reward-based and distillation terms, respectively. 

In detail, consider the objective using unnormalized forward KL regularization as follows: 
\begin{align*}
    J_\mathrm{SDPG-UFKL}(\theta)=J_\mathrm{R\&D}(\theta)-\alpha\mathrm{UKL}(\piR\|\piS),
\end{align*}
where $J_\mathrm{OPD}$ is the objective of on-policy distillation implicitly involved in Eq.~\eqref{eq:opd_loss}. The gradient, expressed as an expectation over $\pi_\theta$ using $w_T(x) = \pi_\theta(x)/\piT(x)$, $w_R = \piS(x)/\piR(x)$, $\piT(x)=\piS(x,c)$ is:
\begin{align*}
\nabla_\theta J_{\mathrm{SDPG-UFKL}}(\theta) = \nabla_\theta J_{\mathrm{R\&D}}(\theta)- \alpha~\mathbb{E}_{x \sim \pi_\theta}\left[(1-w_R(x)^{-1})\nabla_\theta \log \pi_\theta(x) \right].
\end{align*}
A corresponding differentiable surrogate loss term for minimization via gradient descent is (ignoring the prefix $y_{<t}$ already generated):
\begin{align*}
\mathcal{L}_{\mathrm{SDPG-UFKL}}(x, \theta) = \mathcal{L}_{\mathrm{R\&D}}(x, \theta) +\alpha(w_R(x)^{-1}+\log w_R(x)),
\end{align*}
such that $\mathbb{E}_{x \sim \pi_\theta}[\nabla_\theta \mathcal{L}_{\mathrm{SDPG-UFKL}}(x, \theta)] = -\nabla_\theta J_{\mathrm{SDPG-UFKL}}(\theta)$.
Therefore, the total surrogate loss function can be written as 
\begin{align*}
    &\mathcal{L}_\mathrm{SDPG-UFKL}(\theta) = \mathcal{L}_\mathrm{R\&D}(\theta)\nonumber\\&\qquad+\alpha~\mathbb{E}_{(x, c) \sim \mathcal{D}, y \sim \pi_\theta(\cdot \mid x)} \bigg[ \frac{1}{\sum_{i=1}^G|y_i|}\sum_{i=1}^G \sum_{t=1}^{|y_i|}\bigg(\frac{ \piR(y_{i,t} \mid x, y_{i,<t})}{\pi_\theta(y_{i,t} \mid x, y_{i,<t})}+\log\frac{ \piS(y_{i,t} \mid x, y_{i,<t})}{\piR(y_{i,t} \mid x, y_{i,<t})}\bigg)   \bigg].
\end{align*}

Moreover, we can also apply the unnormalized reverse KL regularization as follows: 
\begin{align*}
    J_\mathrm{SDPG-URKL}(\theta)=J_\mathrm{R\&D}(\theta)-\alpha\mathrm{UKL}(\piS\|\piR).
\end{align*}
The gradient, expressed as an expectation over $\pi_\theta$ is:
\begin{align*}
\nabla_\theta J_{\mathrm{SDPG-URKL}}(\theta) = \nabla_\theta J_{\mathrm{R\&D}}(\theta)- \alpha~\mathbb{E}_{x \sim \pi_\theta}\left[(\log w_R(x))\nabla_\theta \log \pi_\theta(x) \right].
\end{align*}
A corresponding differentiable surrogate loss term for minimization via gradient descent is (ignoring the prefix $y_{<t}$ already generated):
\begin{align*}
\mathcal{L}_{\mathrm{SDPG-URKL}}(x, \theta) = \mathcal{L}_{\mathrm{R\&D}}(x, \theta) +\frac{\alpha}{2}\log^2 w_R(x),
\end{align*}
such that $\mathbb{E}_{x \sim \pi_\theta}[\nabla_\theta \mathcal{L}_{\mathrm{SDPG-URKL}}(x, \theta)] = -\nabla_\theta J_{\mathrm{SDPG-URKL}}(\theta)$.
And the total surrogate loss function can be written as 
\begin{align*}
    \mathcal{L}_\mathrm{SDPG-URKL}(\theta) &= \mathcal{L}_\mathrm{R\&D}(\theta)+\alpha~\mathbb{E}_{(x, c) \sim \mathcal{D}, y \sim \pi_\theta(\cdot \mid x)} \bigg[ \frac{1}{\sum_{i=1}^G|y_i|}\sum_{i=1}^G \sum_{t=1}^{|y_i|} \frac{1}{2} \log^2 \frac{ \piS(y_{i,t} \mid x, y_{i,<t})}{\piR(y_{i,t} \mid x, y_{i,<t})}   \bigg].
\end{align*}

\subsection{Additional Stabilizers for SDPG}
\label{sec:sdpg_stabilizers}
Because the current model induces the privileged OPD target under a richer context, it can be noisy early in training and over-constrained late in training. We therefore add two lightweight controls to prevent the privileged OPD signal from overwhelming verifier-grounded learning, including the positive advantage gating and a warmup-then-decay scheduler for the distillation term. These choices trust privileged distillation only on verifier-endorsed rollouts and phase out the privileged signal near the end of training. 

\subsubsection{Positive Advantage Gating}
\label{appendix:pos_gate}

When a rollout is incorrect ($A^{(i)}_{\mathrm{out}}<0$), the privileged teacher can still assign high probability to locally plausible tokens on the sampled wrong prefix. Applying full-vocabulary OPD on such prefixes may conflict with the verifier signal: the outcome objective suppresses the trajectory, whereas the distillation objective can still imitate the privileged teacher around that prefix.

We therefore gate the OPD loss by the outcome advantage:
\begin{align}
  m_i
  =
  \mathbf 1[A^{(i)}_{\mathrm{out}}>0],
  \qquad
  \mathcal L_{\mathrm{OPD}}^{+}(\theta)
  =
  \E_{(x,c)\sim\mathcal D,\;\{y_i\}_{i=1}^{G}\sim\piS(\cdot\mid x)}
  \left[
  \frac{1}{\sum_{i=1}^{G} |y_i|}
  \sum_{i=1}^{G}
  \sum_{t=1}^{|y_i|}
  m_i
  \ell^{\mathrm{OPD}}_{i,t}
  \right].
  \label{eq:advantage_gate}
\end{align}
This relies on the full-vocabulary OPD signal only for trajectories that the verifier endorses within the group. If all rewards in a group are identical, both the mean-centered outcome advantage and the OPD gate vanish, avoiding unvalidated distillation on uninformative groups. In the initial stage, the gate may often be inactive, and the binary outcome reward dominates. A training dataset with moderate difficulty or curriculum learning~\citep{wang2021survey,lee2024instruction,wen2025light,shi2025efficient} is therefore useful for activating the distillation signal. If $m_i=1$ for all responses, Eq.~\eqref{eq:advantage_gate} reduces to standard full-vocabulary OPD on all sampled prefixes.

\subsubsection{\texorpdfstring{$\beta$}{beta} Scheduler}
\label{appendix:beta_schedule}

\begin{wrapfigure}{r}{0.4\textwidth} 
    \centering
    \includegraphics[width=0.4\textwidth]{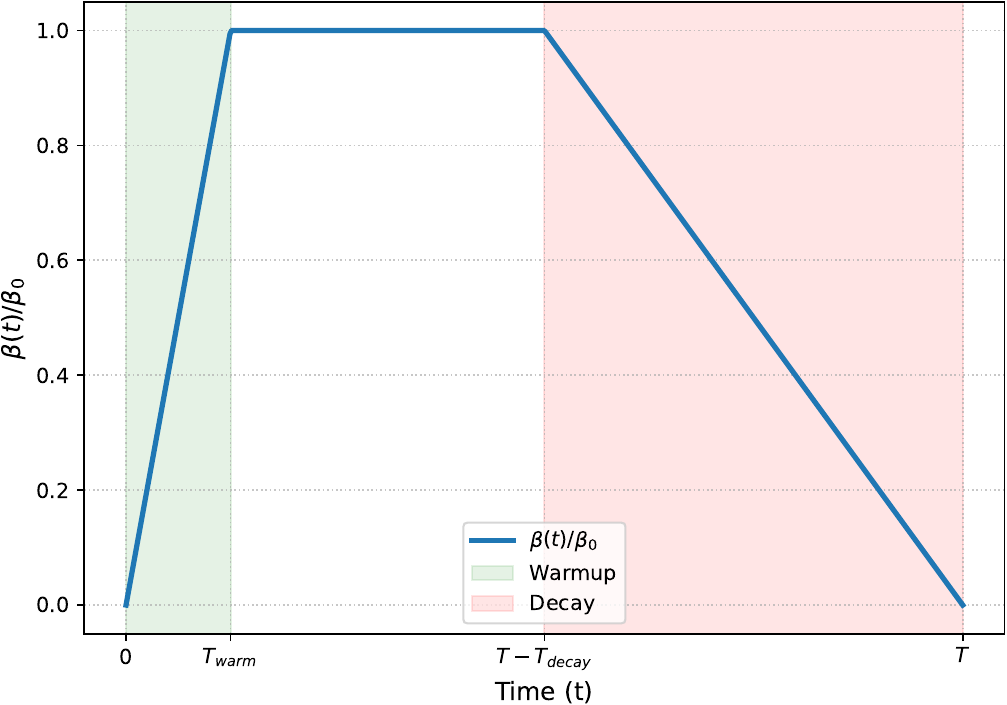}
    \caption{The illustration of the $\beta$ schedule.}
    \label{fig:beta_schedule}
    \vspace{-2em}
\end{wrapfigure}
Early misalignment between $p_t$ and the privileged distribution $q_t$ can make the OPD target noisy. To prevent privileged distillation from destabilizing exploration, we warm up $\beta$. The OPD term then takes effect gradually after the outcome policy has begun to find correct trajectories.

Moreover, under an idealized privileged-information model, distilling a teacher conditioned on information unavailable to the deployable student can leave an irreducible conditional mutual-information gap, e.g., $I(Y_t;C\mid X,Y_{<t})>0$ when $C$ denotes the privileged variable~\citep{yang2026self}. Under our formulation, this means the privileged OPD target may remain biased by information unavailable at inference. Therefore, to release the student and encourage exploration, we decay $\beta$ at the end of training, phasing out the distillation signal after the student has internalized its useful information.

The effective distillation coefficient follows a warmup-decay schedule, illustrated in Figure~\ref{fig:beta_schedule}:
\begin{align*}
  \beta(k)
  =
  \beta_{\mathrm{base}}
  \times
  \underbrace{\min\left(1,\frac{k}{T_{\mathrm{warm}}}\right)}_{\text{warmup}}
  \times
  \underbrace{\min\left(1,\frac{T-k}{T_{\mathrm{decay}}}\right)}_{\text{decay}},
\end{align*}
where $T_{\mathrm{warm}}$ and $T_{\mathrm{decay}}$ are the warmup and decay step counts, and $T$ is the total number of training steps. If the warmup and decay windows overlap, the maximum coefficient can be below $\beta_{\mathrm{base}}$.

\section{Experiments}

In this section, we empirically evaluate our proposed SDPG algorithm and compare the performance against baselines on challenging mathematical reasoning tasks based on pretrained LLMs, including GRPO~\citep{shao2024deepseekmath} and RLSD~\citep{yang2026self}.

\begin{figure}[t]
    \centering
    \includegraphics[width=0.75\linewidth]{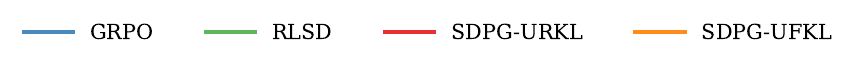}\\[4pt]
    
    \begin{minipage}[t]{0.31\linewidth}
        \centering
        \includegraphics[width=\linewidth]{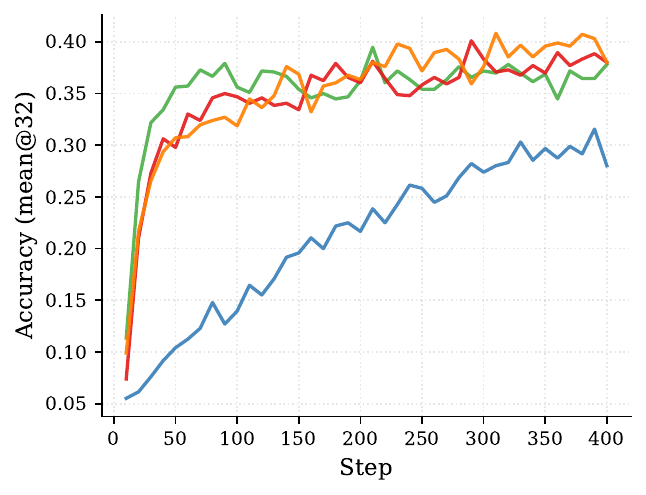}\\
        {\small (a) AIME24}
    \end{minipage}\hfill
    \begin{minipage}[t]{0.31\linewidth}
        \centering
        \includegraphics[width=\linewidth]{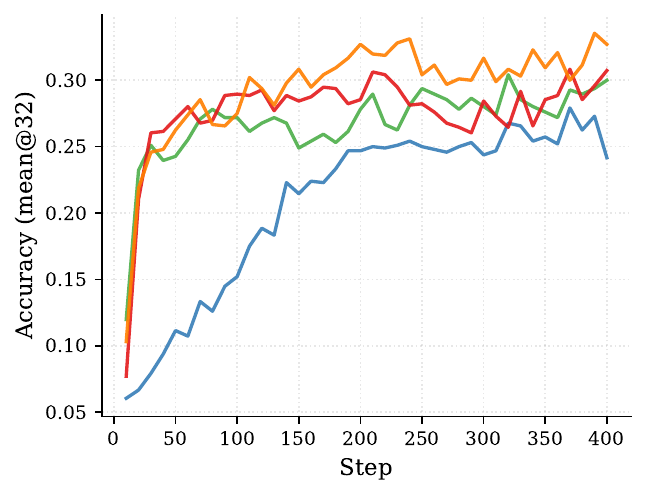}\\
        {\small (b) AIME25}
    \end{minipage}\hfill
    \begin{minipage}[t]{0.31\linewidth}
        \centering
        \includegraphics[width=\linewidth]{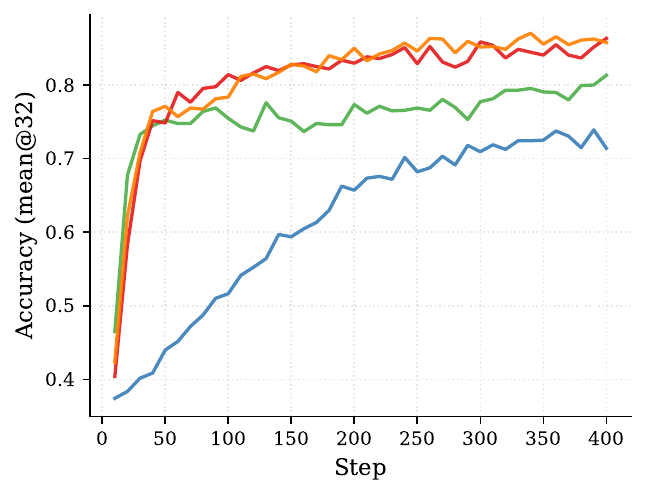}\\
        {\small (c) AMC23}
    \end{minipage}\\[8pt]
    
    \begin{minipage}[t]{0.31\linewidth}
        \centering
        \includegraphics[width=\linewidth]{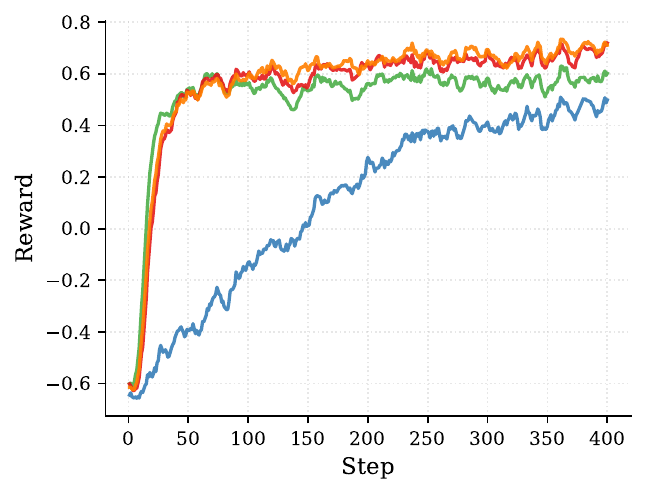}\\
        {\small (d) Reward}
    \end{minipage}\hfill
    \begin{minipage}[t]{0.31\linewidth}
        \centering
        \includegraphics[width=\linewidth]{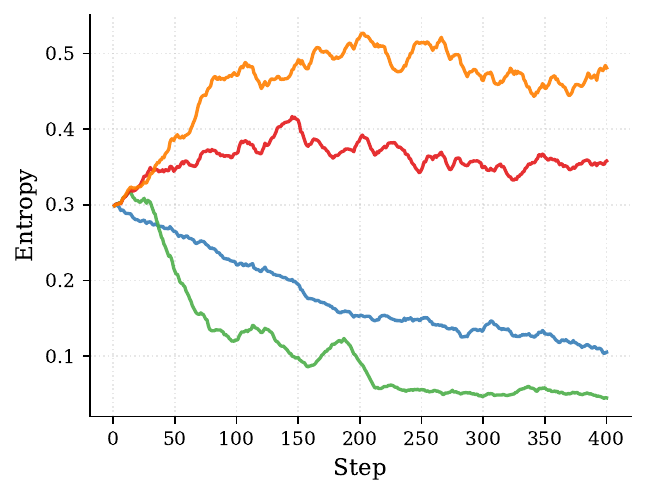}\\
        {\small (e) Entropy}
    \end{minipage}\hfill
    \begin{minipage}[t]{0.31\linewidth}
        \centering
        \includegraphics[width=\linewidth]{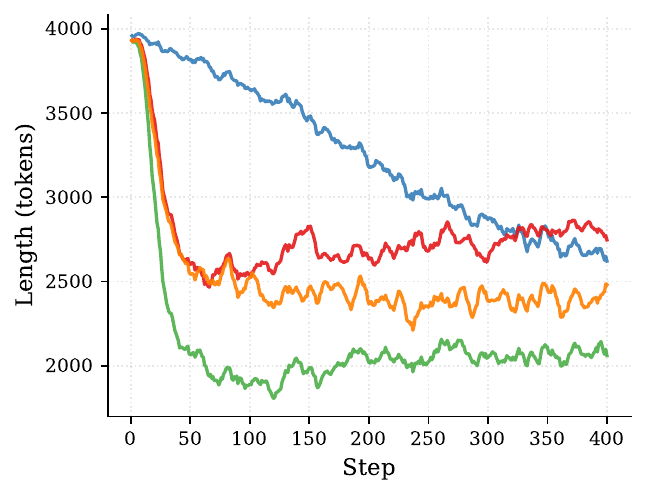}\\
        {\small (f) Response Length}
    \end{minipage}
    
    \caption{Training dynamics and benchmark performance on Qwen3-4B trained with baseline algorithms and SDPG variants. \textbf{Top row:} (a) AIME24, (b) AIME25, (c) AMC23. \textbf{Bottom row:} (d) group-relative reward, (e) actor entropy, (f) average response length. Both SDPG-URKL and SDPG-UFKL reach higher final accuracies than baselines. SDPG-UFKL also avoids the entropy collapse observed in RLSD.}
    \label{fig:main_4b}
\end{figure}

\subsection{Experiment Settings}

\begin{figure}[htbp]
\centering
\begin{tikzpicture}[
    box/.style={
        rectangle,
        rounded corners=6pt,
        draw=#1!70!black,
        line width=0.6pt,
        fill=#1!8,
        drop shadow={shadow xshift=1.5pt, shadow yshift=-1.5pt, opacity=0.25},
        text width=12cm,
        align=left,
        inner sep=10pt,
    },
    title/.style={
        rectangle,
        rounded corners=4pt,
        fill=#1!75!black,
        text=white,
        font=\bfseries\sffamily,
        inner sep=5pt,
        minimum width=11.6cm,
        align=center,
    },
    content/.style={
        font=\small\sffamily,
        text width=11.6cm,
        align=left,
    }
]

\node[box=blue] (student) at (0,0) {
    \\[2em]
    Solve the following math problem step by step. Present your final answer inside $\backslash$boxed\{\}, for example $\backslash$boxed\{42\}. \\[4pt]
    \{question\} \\[4pt]
    Remember to put your final answer inside $\backslash$boxed\{\}.
};
\node[title=blue, anchor=north] at ([yshift=-8pt]student.north) {Student Prompt};

\node[box=orange, below=0.6cm of student] (teacher) {
    \\[2em]
    Solve the following math problem step by step. Present your final answer inside $\backslash$boxed\{\}, for example $\backslash$boxed\{42\}. \\[4pt]
    \{question\} \\[4pt]
    Remember to put your final answer inside $\backslash$boxed\{\}. [TEACHER\_CONTEXT\_TOKEN] \\[4pt]
    [Hint] The correct answer is \{answer\}. A common way to solve this is: \{solution\} \\[4pt]
    [Instruction] If possible, derive the answer \{answer\} using an alternative, equally rigorous mathematical approach (e.g., algebraic vs geometric, or different substitution). If no alternative exists, articulate the standard approach with exceptional clarity. Do NOT state that you were given the answer or reference.
};
\node[title=orange, anchor=north] at ([yshift=-8pt]teacher.north) {Teacher Prompt};

\end{tikzpicture}
\caption{Prompt templates for the student and teacher models, where the ``\{question\}'', ``\{answer\}'' are from the dataset and the ``\{solution\}'' is generated by Gemini 2.5 Pro.}
\label{fig:prompts}
\end{figure}
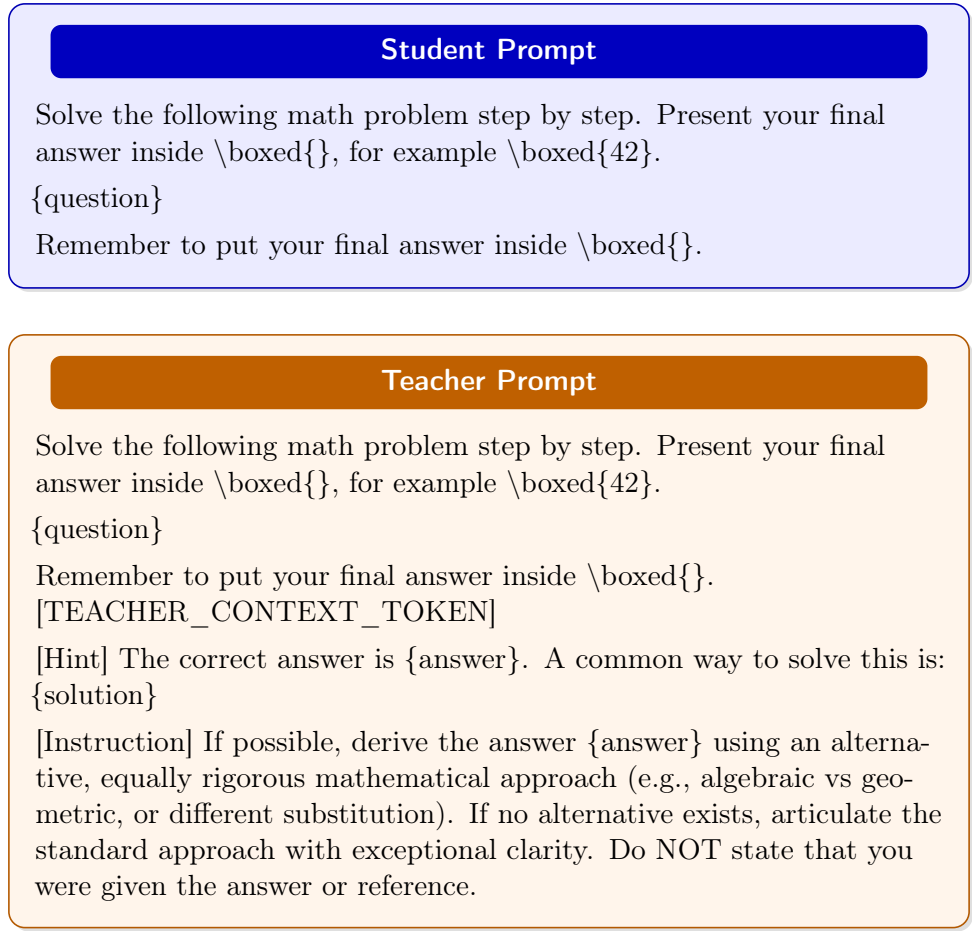

We conduct experiments primarily on Qwen3-4B~\citep{yang2025qwen3}; additional results on Qwen3-1.7B with the further baseline, OPCD, are reported in Appendix~\ref{appendix:ablation_1p7b}. For training, we utilize the DAPO-Math-17k dataset~\citep{yu2025dapo} with 13.9k English samples and generate the privileged information using Gemini 2.5 Pro~\citep{gemini25report2025}. And the prompts for teacher and student models are shown in Figure~\ref{fig:prompts}. We evaluate the fine-tuned models on the benchmarks of AIME2024~\citep{AIME2024I,AIME2024II}, AIME2025~\citep{AIME2025I,AIME2025II}, and AMC23~\citep{AMC23}. Experiments are implemented using the verl framework~\citep{sheng2025hybridflow} with the vLLM engine~\citep{kwon2025vllm} for efficient LLM serving and inference.

All experiments use the AdamW optimizer~\citep{loshchilov2018decoupled} with a learning rate of $1\times10^{-6}$, a weight decay of $0.1$ $(\beta_1,\beta_2)=(0.9,0.999)$, and gradient clipping at $1.0$. Training proceeds for 400 steps with 10 warmup steps in the beginning. The global training batch size is 128, with 8 responses per prompt and a temperature of $1.0$. We use FSDP with bfloat16 mixed precision and the vLLM rollout engine, and all experiments are conducted on 8 NVIDIA H100 GPUs. The maximum prompt and response lengths are set to 2,048 and 4,096, respectively, with dynamic batching enabled. For all the baselines, we use $\epsilon_\mathrm{std}=1e-6$ in Eq.~\eqref{eq:grpo_advantage}, and use a clipping threshold of $(\epsilon_1,\epsilon_2)=(0.2,0.2)$, and the KL regularization coefficient is set to $1\times10^{-3}$. For SDPG, we use $\alpha=1\times10^{-3}$, $\beta_{\mathrm{base}}=1\times10^{-3}$, $T_{\mathrm{warm}}=50$, and $T_{\mathrm{decay}}=350$.

\subsection{Experiment Results}
\label{sec:exp_results}

The quantitative results in Table~\ref{tab:main_4b} demonstrate the competitive performance of the proposed SDPG framework. Both SDPG-URKL and SDPG-UFKL outperform GRPO and RLSD across all the benchmarks, with SDPG-UFKL achieving the top score in five and SDPG-URKL in the remaining one. Figure~\ref{fig:main_4b} complements these results by illustrating evaluation scores and training dynamics. The accuracy gap between SDPG and GRPO opens within the first 50 steps and persists throughout training (Figures~\ref{fig:main_4b}a--c), and SDPG reaches the high-reward plateau (Figure~\ref{fig:main_4b}d) several hundred steps earlier than GRPO. Notably, SDPG-UFKL maintains substantially higher actor entropy throughout training (Figure~\ref{fig:main_4b}e), in contrast to RLSD whose entropy collapses toward zero by step 250, a known signature of mode collapse in pure self-distillation. We attribute this stability to the combination of positive-advantage gating and the warmup-decay $\beta$ schedule, which together prevent privileged distillation from over-constraining the policy after the student has internalized its useful signal. Response lengths (Figure~\ref{fig:main_4b}f) for SDPG methods stabilize at intermediate values, sufficient for multi-step reasoning while remaining shorter than GRPO's verbose outputs. Ablation studies isolating the contributions of the OPD distillation term and the KL regularization, as well as additional results on Qwen3-1.7B, are reported in Appendix~\ref{appendix:ablation}.

\begin{table}[t]
\centering
\caption{Performance of models trained with baselines and SDPG variants on AIME24, AIME25, and AMC23 (pass@1, mean@32) with Qwen3-4B trained for 400 steps. \emph{Last} shows the score at step 400; \emph{Best} shows the peak across training. Column maximum in \textbf{bold}, second-best in \underline{underlined}.}
\label{tab:main_4b}
\small
\begin{tabular}{l cc cc cc}
\toprule
& \multicolumn{2}{c}{\textbf{AIME24}} & \multicolumn{2}{c}{\textbf{AIME25}} & \multicolumn{2}{c}{\textbf{AMC23}} \\
\cmidrule(lr){2-3} \cmidrule(lr){4-5} \cmidrule(lr){6-7}
Method & Last & Best & Last & Best & Last & Best \\
\midrule
GRPO              & 0.280 & 0.316 & 0.242 & 0.279 & 0.714 & 0.739 \\
RLSD              & 0.378 & 0.395 & 0.300 & 0.304 & 0.813 & 0.813 \\
\midrule
SDPG-URKL (ours) & \underline{0.380} & \underline{0.401} & \underline{0.307} & \underline{0.308} & \textbf{0.863} & \underline{0.863} \\
SDPG-UFKL (ours) & \textbf{0.380} & \textbf{0.408} & \textbf{0.327} & \textbf{0.335} & \underline{0.858} & \textbf{0.870} \\
\bottomrule
\end{tabular}
\end{table}

\section{Related Work}

\paragraph{Reinforcement Learning with Verifiable Rewards.}
Reinforcement Learning with Verifiable Rewards (RLVR)~\citep{wang2025reinforcement} has become one of the dominant techniques in post-training stages of LLMs, which reward models with an automatic verifier for achieving correct final answers regardless of intermediate reasoning steps~\citep{srivastava2025technical}. It eliminates the need for human preference labels in previous algorithms like PPO~\citep{schulman2017proximal}. One classic paradigm is GRPO~\citep{shao2024deepseekmath}, which utilizes group-relative advantage, PPO-style clipped surrogate, as well as KL regularization to achieve competitive performances when training LLMs. However, GRPO has limitations: advantages collapse and learning stalls when all rollouts in the group have identical rewards, and token-wise credit assignment remains sparse in long generations. Therefore, a lot of improved algorithms have been proposed. Dr. GRPO~\citep{liu2025understanding} removes biased normalization to improve token efficiency. VinePPO~\citep{kazemnejad2024vineppo} utilizes Monte-Carlo-based value estimation to improve credit assignment. Moreover, DAPO~\citep{yu2025dapo} introduces techniques including clip-higher, dynamic sampling, token-level loss, and soft overlong punishment to enhance downstream performance, while GSPO~\citep{zheng2025group} uses sequence likelihood in the importance ratio to further stabilize training. RPG~\citep{zhang2026design} provides a unified framework for different types of KL regularization in GRPO.

However, these approaches may suffer from signal homogeneity among all tokens in a sequence. Some research utilizes process reward models~\citep{lightman2023let,wang2024math,chen2024step,zhang2025generative,dai2025s,yang2026test,luo2024improve} or step-level value estimators for intermediate signals. However, expensive human step annotations may make these methods less affordable. Beyond these, other works try to use per-token signals such as entropy, key-token statistics, uncertainty, or attention dynamics to circumvent expensive annotations~\citep{xie2025unlocking,li2026outcome,cheng2026reasoning,wang2025beyond,chen2025seed}. However, such utilization of intrinsic signals may be heuristic. Our method instead incorporates full-vocabulary privileged self-distillation while retaining the binary outcome verifier.

\paragraph{On-Policy Distillation and Self-Distillation.}
To deal with sparse rewards in reinforcement learning, on-policy distillation (OPD) techniques have been proposed to provide dense token-level supervision signals~\citep{agarwal2024policy,lu2025onpolicy,fu2026revisiting}. They train a student model based on trajectories sampled from its own policy, while another teacher model provides token-level targets via KL regularization or related objectives~\citep{gu2024minillm,xu2025speculative,yang2025qwen3,xiao2026mimo}. However, these OPD methods often require a large external teacher, which requires much larger memory usage and may provide mismatched guidance due to model heterogeneity.

To address such issues, self-distillation methods sample from the same student policy but evaluate the model under privileged knowledge, including ground-truth solution paths and environmental feedback~\citep{hubotter2026reinforcement,shenfeld2026self,penaloza2026privileged}. In detail, OPSD~\citep{zhao2026self} utilizes full KL divergence between teacher and student models to enhance the reasoning ability of LLMs; OPCD~\citep{ye2026policy} adds flexibility by decoupling the on-policy strategy; and TRRD~\citep{zhang2026reinforcement} involves the teacher policy in the importance ratio to alleviate the conflict between reward function and distillation term. However, pure on-policy self-distillation approaches suffer from limited exploration and mode collapse. RLSD~\citep{yang2026self} instead uses the privileged teacher-student likelihood ratio as a token-level credit reweighting signal inside a GRPO-style objective; we summarize this distinction in Section~\ref{appendix:rlsd_details} and provide the full loss in Appendix~\ref{appendix:rlsd_details}. By taking advantage of RLVR techniques, the training process becomes smoother. However, strong teacher signals can still impose a lower ceiling on the student~\citep{xiao2026mimo}. SDPG differs from these methods by preserving the exact full-vocabulary OPD objective while coupling it with verifier-based policy optimization and reference-policy KL regularizationing.

\section{Conclusion}

We presented SDPG, a Self-Distilled Policy Gradient framework that combines verifier-based RLVR with exact full-vocabulary OPD. In this view, the reverse-KL OPD term remains a full-vocabulary distillation objective, while its fixed-prefix student-side gradient admits an equivalent policy-gradient form with a centered log-ratio token advantage. Combining this OPD-equivalent distillation signal with binary verifier rewards improves credit assignment while retaining the exploration and selection benefits of RLVR. On LLM reasoning tasks, the proposed algorithms achieve better performance and stability than baseline algorithms.

\section*{Acknowledgement}
Thank Fetch Compute program for their support of compute resources.

\bibliography{reference}
\bibliographystyle{ims}


\newpage
\appendix
\renewcommand{\appendixpagename}{\centering \huge Appendix}
\appendixpage
\counterwithin{theorem}{section}

\startcontents[section]
\printcontents[section]{l}{1}{\setcounter{tocdepth}{2}}
\clearpage

\section{More on SDPG Loss}
\label{appendix:deferred_sdpg_derivations}

\subsection{Proof of Proposition~\ref{prop:opd_advantage}}
\label{sec:opd_equivalent_policy_gradient}
For a rollout prefix $s_t=(x,y_{<t})$, recall the unprivileged student distribution and the privileged teacher distribution:
\begin{align*}
    p_t(a)&=\piS(a\mid s_t)=\pi_\theta(a\mid x,y_{<t}),\\
    q_t(a)&=\piT(a\mid s_t)=\pi_\theta(a\mid c,x,y_{<t}).
\end{align*}
All identities here are local to a fixed prefix $s_t$ and to the current iterate used to form the detached quantities. Write
\[
    \bar p_t=\SG[p_t],
    \qquad
    \bar q_t=\SG[q_t],
    \qquad
    \bar D_t=\KL(\bar p_t\|\bar q_t).
\]
The sampled-prefix distribution, the teacher branch, and any policy-gradient coefficients are treated as detached, as in standard surrogate optimization. Reverse-KL full-vocabulary on-policy distillation minimizes
\begin{align}
    \mathcal L_{\mathrm{OPD},t}(\theta)
    =
    \KL(p_t\|\bar q_t)
    =
    \sum_{a\in\mathcal V}
    p_t(a)
    \log
    \frac{p_t(a)}{\bar q_t(a)}.
    \label{eq:appendix_opd_kl}
\end{align}
The negative student-side gradient of Eq.~\eqref{eq:appendix_opd_kl} is
\begin{align*}
    -\nabla_\theta \mathcal L_{\mathrm{OPD},t}(\theta)
    &=
    -\sum_{a\in\mathcal V}
    p_t(a)\left(\log\frac{p_t(a)}{\bar q_t(a)} + 1\right)
    \nabla_\theta \log p_t(a)\nonumber\\
    &=
    \E_{a\sim \bar p_t}
    \left[
        -\frac{p_t(a)}{\bar p_t(a)}
        \left(\log\frac{p_t(a)}{\bar q_t(a)}+1\right)
        \nabla_\theta \log p_t(a)
    \right],
\end{align*}
where gradients are evaluated at the iterate satisfying $\bar p_t=p_t$.
At the iterate where $\bar p_t=p_t$,
\[
    \E_{a\sim \bar p_t}[\nabla_\theta\log p_t(a)]
    =
    \sum_{a\in\mathcal V}p_t(a)\nabla_\theta\log p_t(a)
    =
    \nabla_\theta\sum_{a\in\mathcal V}p_t(a)
    =0.
\]
Adding the state-dependent baseline $1+\bar D_t$ therefore leaves the gradient unchanged and yields
\begin{align*}
    -\nabla_\theta \mathcal L_{\mathrm{OPD},t}(\theta)
    =
    \E_{a\sim \bar p_t}
    \left[
        \left(\bar D_t-\log\frac{\bar p_t(a)}{\bar q_t(a)}\right)
        \nabla_\theta\log p_t(a)
    \right],
\end{align*}
which is the negative gradient of the detached-sampling surrogate in Eq.~\eqref{eq:opd_advantage_pg}. Centering follows from
\[
    \E_{a\sim \bar p_t}[\Adist_t(a)]
    =
    \bar D_t
    -
    \sum_{a\in\mathcal V}\bar p_t(a)\log\frac{\bar p_t(a)}{\bar q_t(a)}
    =
    0.
\]
\hfill$\square$

Thus, on a fixed sampled prefix and with the teacher branch detached, reverse-KL full-vocabulary OPD has a local policy-gradient interpretation with centered log-ratio advantage. SDPG nevertheless implements the explicit full-vocabulary KL in Eq.~\eqref{eq:appendix_opd_kl}; the policy-gradient form is an interpretation of its gradient, not a sampled-token replacement. The same identity also yields sampled-token Monte Carlo estimators by replacing the full-vocabulary expectation with samples from the detached student distribution.

\subsection{Normalized KL terms}
\label{sec:normalized_KL}
In common implementations such as GRPO~\citep{shao2024deepseekmath}, a forward or reverse KL term is directly applied to the final loss functions. However, if such KL terms are only rollout-based estimation, they could be biased and here we derive the correct loss functions for these KL regularizations. 

Firstly, we consider the objective for forward KL regularization as follows: 
\begin{align*}
    J_\mathrm{FKL}=J_\mathrm{R\&D}-\alpha\KL(\piR\|\piS).
\end{align*}
The gradient, expressed as an expectation over $\pi_\theta$ using $w_T(x) = \pi_\theta(x)/\piT(x)$, $w_R = \piS(x)/\piR(x)$, $\piT(x)=\piS(x,c)$ is:
\begin{align*}
\nabla_\theta J_{\mathrm{FKL}}(\theta) = \nabla_\theta J_{\mathrm{R\&D}}(\theta)+\alpha~\mathbb{E}_{x \sim \pi_\theta}[ w_R(x)^{-1} \nabla_\theta \log \pi_\theta(x)].
\end{align*}
A corresponding differentiable surrogate loss term for minimization via gradient descent is (ignoring the prefix $y_{<t}$ already generated):
\begin{align*}
\mathcal{L}_{\mathrm{FKL}}(x, \theta) = \mathcal{L}_{\mathrm{R\&D}}(x, \theta) +\alpha w_R(x)^{-1},
\end{align*}
such that $\mathbb{E}_{x \sim \pi_\theta}[\nabla_\theta \mathcal{L}_{\mathrm{FKL}}(x, \theta)] = -\nabla_\theta J_{\mathrm{FKL}}(\theta)$.
And the total surrogate loss function can be written as 
\begin{align*}
    \mathcal{L}_\mathrm{FKL}(\theta) &= \mathcal{L}_\mathrm{R\&D}(\theta)+\alpha~\mathbb{E}_{(x, c) \sim \mathcal{D}, y \sim \pi_\theta(\cdot \mid x)} \bigg[ \frac{1}{\sum_{i=1}^G|y_i|}\sum_{i=1}^G \sum_{t=1}^{|y_i|}\frac{ \piR(y_{i,t} \mid x, y_{i,<t})}{\pi_\theta(y_{i,t} \mid x, y_{i,<t})}   \bigg]
\end{align*}

For the reverse KL regularization, consider the objective as follows: 
\begin{align*}
    J_\mathrm{RKL}=J_\mathrm{R\&D}-\alpha\KL(\piS\|\piR).
\end{align*}
The gradient, expressed as an expectation over $\pi_\theta$ using $w_T(x) = \pi_\theta(x)/\piT(x)$, $w_R = \piS(x)/\piR(x)$, $\piT(x)=\piS(x,c)$ is:
\begin{align*}
\nabla_\theta J_{\mathrm{RKL}}(\theta) = \nabla_\theta J_{\mathrm{R\&D}}(\theta)- \alpha~\mathbb{E}_{x \sim \pi_\theta}[(\log w_R(x)+1) \nabla_\theta \log \pi_\theta(x)].
\end{align*}
A corresponding differentiable surrogate loss term for minimization via gradient descent is (ignoring the prefix $y_{<t}$ already generated):
\begin{align*}
\mathcal{L}_{\mathrm{RKL}}(x, \theta) = \mathcal{L}_{\mathrm{R\&D}}(x, \theta) +\frac{\alpha}{2} (\log w_R(x)+1)^2,
\end{align*}
such that $\mathbb{E}_{x \sim \pi_\theta}[\nabla_\theta \mathcal{L}_{\mathrm{RKL}}(x, \theta)] = -\nabla_\theta J_{\mathrm{RKL}}(\theta)$.
And the total surrogate loss function can be written as 
\begin{align*}
    \mathcal{L}_\mathrm{RKL}(\theta) &= \mathcal{L}_\mathrm{R\&D}(\theta)+\alpha~\mathbb{E}_{(x, c) \sim \mathcal{D}, y \sim \pi_\theta(\cdot \mid x)} \bigg[ \frac{1}{\sum_{i=1}^G|y_i|}\sum_{i=1}^G \sum_{t=1}^{|y_i|} \frac{1}{2}\bigg( 1+\log \frac{ \pi_\theta(y_{i,t} \mid x, y_{i,<t})}{\piR(y_{i,t} \mid x, y_{i,<t})}\bigg)^2   \bigg].
\end{align*}

It can be observed that due to expectation over $\pi_\theta$, which is differentiable, although both gradient and expectation operations are linear-operator, they cannot commute and the gradient of expectation is not the same as the expectation of the gradient. Therefore, for rollout-based KL regularization, the original KL loss forms are actually biased.

\subsection{Analysis in one-step off-policy settings}
In the implementation of modern RL frameworks, such as verl~\citep{sheng2025hybridflow}, the rollout model $\pi_\mathrm{rollout}$ is often different from the current model $\pi_\theta$. For example, there would be some sub-iterations of gradient updates for mini-batches in each step. Therefore, there could be issues of within-step off-policy drift and stale importance weights. Therefore, the loss would be different from the full on-policy implementations, and an importance sampling ratio factor should be applied in the loss function. Based on Eq.~\ref{eq:SDPG_loss}, we have

\begin{align*}
\mathcal L_{\mathrm{SDPG}}(\theta)&=
\mathcal L_{\mathrm{out}}(\theta)
+
\beta(k)\mathcal L_{\mathrm{OPD}}(\theta)
+
\alpha\mathcal L_{\mathcal K}(\pi_\theta,\piR)\nonumber\\
&=
\E_{(x,c)\sim\mathcal D,\;\{y_i\}_{i=1}^{G}\sim\piS(\cdot\mid x)}
\bigg[
\frac{1}{\sum_{i=1}^{G}|y_i|}
\sum_{i=1}^{G}
\sum_{t=1}^{|y_i|}
-\SG[A_\mathrm{out}(y_i)]\,
\log \pi_\theta(y_{i,t}\mid x,y_{i,<t})\nonumber\\
&\qquad+\beta \mathcal{L}_\mathrm{OPD}(x,\theta)+\alpha f_\mathcal{K}(\piS,\pi_R;x,y_i)\bigg]\nonumber\\
&=
\E_{(x,c)\sim\mathcal D,\;\{y_i\}_{i=1}^{G}\sim\pi_\mathrm{rollout}(\cdot\mid x)}
\bigg[
\frac{1}{\sum_{i=1}^{G}|y_i|}
\sum_{i=1}^{G}
\sum_{t=1}^{|y_i|}
-\SG[\rho_{i,t}A_i]\,
\log \pi_\theta(y_{i,t}\mid x,y_{i,<t})\nonumber\\
&\qquad+\beta \SG[\rho_{i,t}]\mathcal{L}_\mathrm{OPD}(x,\theta)+\alpha \SG[\rho_{i,t}]f_\mathcal{K}(\piS,\pi_R;x,y_i)\bigg],
\end{align*}
where
\begin{align*}
    \rho_{i,t}=\rho(y_{i,t}\mid x,y_{i,<t})=\frac{\piS(y_{i,t}\mid x,y_{i,<t})}{\pi_\mathrm{rollout}(y_{i,t}\mid x,y_{i,<t})}
\end{align*}
is the importance ratio during sampling and $A_i=A_\mathrm{out}(y_i)$. And $f_\mathcal{K}$ is the corresponding KL surrogate loss.

For the reward-based term, usually we can apply a PPO-style clip to stabilize the training process. In detail, the clipped advantage with the importance ratio is
\begin{align*}
    \ell_{t,i}^\mathrm{clip}=\begin{cases}
\min\!\bigl(\max(-A_i \rho_{t,i},\,-A_i\,\text{clip}(\rho_{t,i},1{-}\varepsilon_l,1{+}\varepsilon_h)),\;-A_i c\bigr) & A_i < 0 \\
\max(-A_i \rho_{t,i},\,-A_i\,\text{clip}(\rho_{t,i},1{-}\varepsilon_l,1{+}\varepsilon_h)) & A_i \geq 0
\end{cases},
\end{align*}

where $\epsilon_l,\epsilon_h$ as well as $c$ are the clipping hyperparameters. Here we can set $\epsilon_h=\epsilon_l$ as in GRPO~\citep{shao2024deepseekmath} and RLSD~\citep{yang2026self}, or $\epsilon_h>\epsilon_l$ as in DAPO~\citep{yu2025dapo}. Therefore, we achieve the final loss in near on-policy (one-step off-policy) approximation implementation:
\begin{align*}
\mathcal L_{\mathrm{SDPG}}^\mathrm{approx}(\theta)
&=
\E_{(x,c)\sim\mathcal D,\;\{y_i\}_{i=1}^{G}\sim\pi_\mathrm{rollout}(\cdot\mid x)}
\bigg[
\frac{1}{\sum_{i=1}^{G}|y_i|}
\sum_{i=1}^{G}
\sum_{t=1}^{|y_i|}
\SG[\ell_{t,i}^\mathrm{clip}]\,
\log \pi_\theta(y_{i,t}\mid x,y_{i,<t})\nonumber\\
&\qquad+\beta \SG[\rho_{i,t}]\mathcal{L}_\mathrm{OPD}(x,\theta)+\alpha \SG[\rho_{i,t}]f_\mathcal{K}(\piS,\pi_R;x,y_i)\bigg].
\end{align*}

\subsection{Discussion about OPSD-style distillation}

While we use OPSD-style full-vocabulary on-policy distillation term, deploying this objective in modern distributed training frameworks (e.g., FSDP with vLLM rollouts) requires specific implementation. In this section, we provide a rigorous analysis of the exact loss computed in our implementation, analyzing the discrepancies between the theoretical gradients and the practical approximations.

The full OPSD-style loss function is given by:

\[\mathcal{L}_{\mathrm{OPD}}(\theta)
= \mathbb{E}_{\substack{(x,c)\sim\mathcal{D}\\\{y_i\}_{i=1}^G\sim\pi_\theta(\cdot\mid x)}}
\left[\frac{1}{\sum_i|y_i|}
\sum_{i,t}
    \mathbb{E}_{v\sim\pi_\theta(\cdot\mid x,y_{i,<t})}\left[\log\frac{\pi_\theta(v\mid x,y_{i,<t})}{\SG[\pi_\theta(v\mid c,x,y_{i,<t})]}\right]
\right].\]

And the corresponding objective is:

\[J(\theta) = \mathbb{E}_{y \sim \pi_\theta} \left[ \sum_{t} D_{\mathrm{KL}}(\pi_\theta \| \SG[\pi_{\text{teacher}}]) \right],\]

where $\pi_\mathrm{teacher}=\pi_\theta(\cdot|c,\cdot)$. When differentiating this objective, it yields two distinct gradient paths for the distillation term:

\[\nabla_\theta \mathbb{E}_{y \sim \pi_\theta} \left[ \sum_t D_{\mathrm{KL}} \right] = \underbrace{\mathbb{E}_{y} \left[ \nabla_\theta \sum_t D_{\mathrm{KL}} \right]}_{\text{(1) Direct Path-wise Gradient}} + \underbrace{\mathbb{E}_{y} \left[ \sum_t D_{\mathrm{KL}} \cdot \nabla_\theta \log \pi_\theta(y_t) \right]}_{\text{(2) Score Function Gradient}}\]

The direct path-wise gradient term corresponds to $\mathcal{L}_\mathrm{OPD}(\theta)$, which is the actual implementation in most modern RL frameworks. However, the score function term is usually omitted, which actually treats the computed KL divergence as an additional negative reward signal. It is worth noting that the magnitude of this term ($\beta \cdot D_{\mathrm{KL}}$) is negligible compared to the primary sequence-level outcome advantage $A_i$. And the variance introduced by estimating it would outweigh its marginal theoretical benefit. Therefore, the distillation term acts almost as a local shaping constraint as in term (1). Therefore, we employ the OPSD-style  on-policy self-distillation loss as a suitable approximation.

\section{Reinforcement Learning with Self-Distillation}
\label{appendix:rlsd_details}

RLSD~\citep{yang2026self} combines verifier-grounded RLVR with privileged self-distillation, but it uses the privileged model only to redistribute token-level credit rather than to define an auxiliary distribution-matching objective. For a sampled token $y_{i,t}$, RLSD computes a stop-gradient privileged information gain $\Delta_{i,t}=\SG[\log q_{i,t}(y_{i,t})-\log p_{i,t}(y_{i,t})]$, where $q_{i,t}$ is the privileged teacher distribution and $p_{i,t}$ is the deployable student distribution. This gain is exponentiated with the sign of the sequence-level verifier advantage and used to reweight the GRPO token advantage. Consequently, tokens favored by the privileged context receive larger positive credit on successful trajectories, while the ratio is inverted for negative-advantage trajectories.

The key distinction from SDPG is that RLSD does not optimize a separate full-vocabulary distribution-matching OPD loss. The privileged model only changes the magnitude of the verifier-grounded policy-gradient update, whereas SDPG keeps the exact full-vocabulary reverse-KL OPD objective and combines it with outcome-reward optimization. 

For a sampled response $y^{(i)}$ and prefix $(x,y_{i,<t})$, define the deployable student distribution and privileged teacher distribution as
\begin{align*}
    p_{i,t}(a)
    &=
    \pi_\theta(a\mid x,y_{i,<t}),\\
    q_{i,t}(a)
    &=
    \pi_\theta(a\mid c,x,y_{i,<t}).
\end{align*}
RLSD computes the stop-gradient privileged information gain on the sampled token:
\begin{align*}
    \Delta_{i,t}
    =
    \SG\left[
        \log q_{i,t}(y_{i,t})
        -
        \log p_{i,t}(y_{i,t})
    \right].
\end{align*}
The gain is converted into a direction-aware evidence weight by using the sign of the sequence-level verifier advantage:
\begin{align*}
    u_{i,t}
    &=
    \exp\left(\operatorname{sign}(A^{(i)})\Delta_{i,t}\right)\\
    &=
    \left(
        \frac{q_{i,t}(y_{i,t})}{p_{i,t}(y_{i,t})}
    \right)^{\operatorname{sign}(A^{(i)})}.
\end{align*}
The evidence weight is clipped and interpolated with the original GRPO advantage:
\begin{align*}
    \widehat A^{\mathrm{RLSD}}_{i,t}
    =
    A^{(i)}
    \left[
        (1-\lambda_{\mathrm{rlsd}})
        +
        \lambda_{\mathrm{rlsd}}
        \operatorname{clip}(u_{i,t},1-\epsilon_w,1+\epsilon_w)
    \right],
\end{align*}
where $\lambda_{\mathrm{rlsd}}\in[0,1]$ controls the strength of self-distilled credit redistribution and $\epsilon_w$ bounds the per-token credit deviation. Setting $\lambda_{\mathrm{rlsd}}=0$ recovers the uniform GRPO advantage, while $\lambda_{\mathrm{rlsd}}=1$ gives the fully reweighted RLSD advantage.

Under the minimization convention used in this paper, the corresponding GRPO-style RLSD surrogate is
\begin{align*}
\resizebox{1.0\textwidth}{!}{$
\begin{aligned}
    \mathcal L_{\mathrm{RLSD}}(\theta)
    =
    -
    \E_{(x,c)\sim\mathcal D,\;\{y^{(i)}\}_{i=1}^{G}\sim\piO(\cdot\mid x)}
    \left[
    \frac{1}{\sum_{i=1}^{G}|y^{(i)}|}
    \sum_{i=1}^{G}
    \sum_{t=1}^{|y^{(i)}|}
    \min\left(
        \rho_{i,t}\widehat A^{\mathrm{RLSD}}_{i,t},
        \operatorname{clip}(\rho_{i,t},1-\epsilon,1+\epsilon)
        \widehat A^{\mathrm{RLSD}}_{i,t}
    \right)
    \right],
\end{aligned}
$}
\end{align*}
where
\begin{align*}
    \rho_{i,t}
    =
    \frac{
        \pi_\theta(y_{i,t}\mid x,y_{i,<t})
    }{
        \pi_{\mathrm{old}}(y_{i,t}\mid x,y_{i,<t})
    }.
\end{align*}
Equivalently, RLSD replaces the uniform sequence-level GRPO advantage $A^{(i)}$ in Eq.~\eqref{eq:grpo_loss} with the token-dependent advantage $\widehat A^{\mathrm{RLSD}}_{i,t}$. No separate full-vocabulary OPD KL loss is optimized; the privileged teacher affects only the magnitude of token-level credit and not the sign of the verifier-grounded update.

\section{Analysis on On-policy Context Distillation}

Recent advancements like On-Policy Context Distillation (OPCD)~\citep{ye2026policy} attempt to distill in-context knowledge $c$ through on-policy KL matching. In our notation, the student distribution at prefix $s_t=(x,y_{<t})$ is
\[
p_t(a)=\pi_\theta(a\mid x,y_{<t}),
\]
and the privileged distribution is
\[
q_t(a)=\pi_\theta(a\mid c,x,y_{<t}).
\]
Let $\bar q_t=\SG[q_t]$ and $\bar p_t=\SG[p_t]$ denote the detached distributions at the current iterate, and set $\bar D_t=\KL(\bar p_t\|\bar q_t)$. The reverse-KL full-vocabulary OPD objective used by SDPG is
\begin{align*}
    \mathcal L_{\mathrm{OPD},t}
    =
    \KL(p_t\|\bar q_t)
    =
    \sum_{a\in\mathcal V}
    p_t(a)
    \log
    \frac{p_t(a)}{\bar q_t(a)}.
\end{align*}
The corresponding negative student-side gradient at the fixed prefix is
\begin{align*}
    -\nabla_\theta \mathcal L_{\mathrm{OPD},t}
    =
    -\sum_{a\in\mathcal V}
    p_t(a)\left(\log\frac{p_t(a)}{\bar q_t(a)}+1\right)
    \nabla_\theta\log p_t(a).
\end{align*}
Equivalently, at the same fixed prefix and current iterate, the gradient can be written as an expectation over a detached student-token sample:
\[
    -\nabla_\theta \mathcal L_{\mathrm{OPD},t}
    =
    \E_{a\sim \bar p_t}
    \left[
    \SG\left[
        \bar D_t-\log\frac{\bar p_t(a)}{\bar q_t(a)}
    \right]
    \nabla_\theta\log p_t(a)
    \right],
\]
where the added baseline $1+\bar D_t$ is valid because $\bar p_t=p_t$ at the iterate where the surrogate is formed. SDPG uses the explicit full-vocabulary objective $\KL(p_t\|\bar q_t)$ in implementation, so its distillation component is exactly reverse-KL OPD, while the centered log-ratio expression clarifies the corresponding local policy-gradient signal.

However, pure self-distillation differs from SDPG in two important ways. First, it optimizes the privileged teacher signal without a binary outcome verifier, so it has no mechanism to prefer globally correct trajectories over locally plausible but incorrect ones. Second, it usually applies the teacher signal on all trajectories, including trajectories that the verifier would reject. SDPG addresses these issues by combining binary outcome rewards, positive-advantage gating, and reference-policy KL regularizationing.

\section{Ablation Studies}
\label{appendix:ablation}

\subsection{Effect of the Full-Vocabulary OPD Term and KL regularization}
\label{appendix:ablation_terms}

To isolate the individual contributions of the two non-outcome components in the SDPG objective (Eq.~\eqref{eq:SDPG_loss}), we run two ablations on Qwen3-4B: setting $\alpha=0$ removes the policy KL regularization, leaving only the binary outcome reward and full-vocabulary OPD term, and setting $\beta=0$ removes the OPD term, recovering RPG without self-distillation.

\begin{figure}[ht!]
    \centering
    \includegraphics[width=0.75\linewidth]{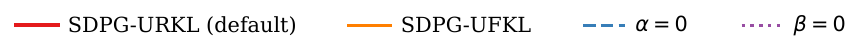}\\[4pt]
    
    \begin{minipage}[t]{0.32\linewidth}
        \centering
        \includegraphics[width=\linewidth]{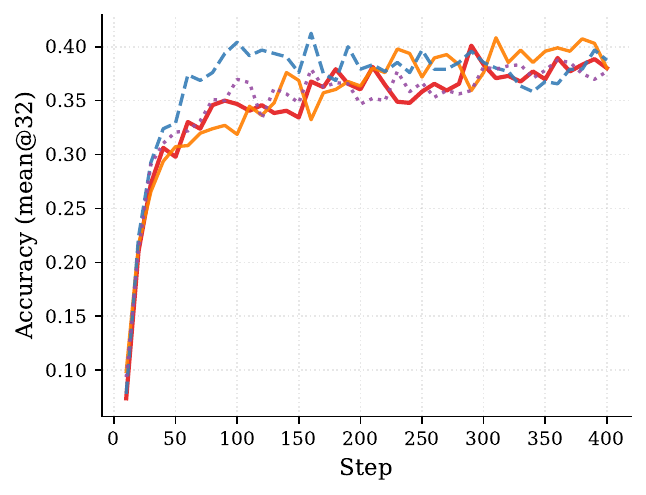}\\
        {\small (a) AIME24}
    \end{minipage}\hfill
    \begin{minipage}[t]{0.32\linewidth}
        \centering
        \includegraphics[width=\linewidth]{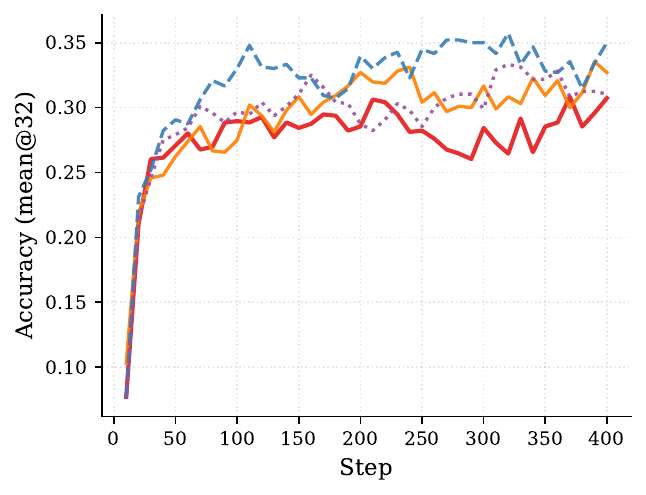}\\
        {\small (b) AIME25}
    \end{minipage}\hfill
    \begin{minipage}[t]{0.32\linewidth}
        \centering
        \includegraphics[width=\linewidth]{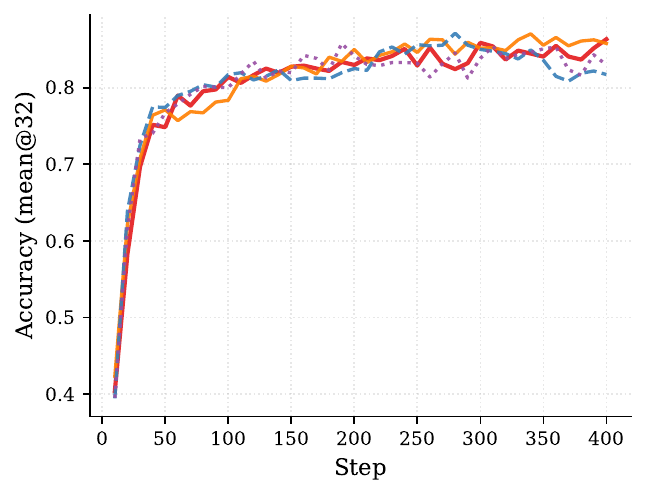}\\
        {\small (c) AMC23}
    \end{minipage}\\[8pt]
    
    \begin{minipage}[t]{0.32\linewidth}
        \centering
        \includegraphics[width=\linewidth]{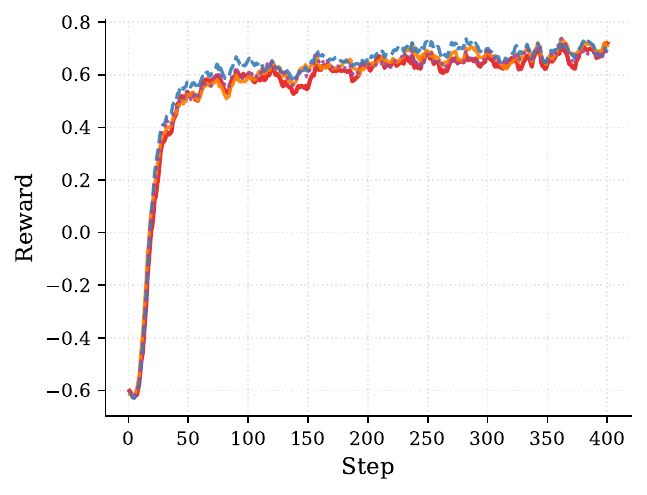}\\
        {\small (d) Reward}
    \end{minipage}\hfill
    \begin{minipage}[t]{0.32\linewidth}
        \centering
        \includegraphics[width=\linewidth]{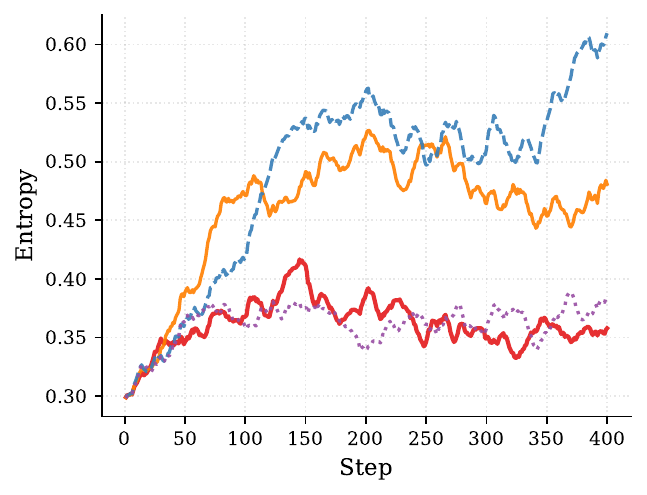}\\
        {\small (e) Entropy}
    \end{minipage}\hfill
    \begin{minipage}[t]{0.32\linewidth}
        \centering
        \includegraphics[width=\linewidth]{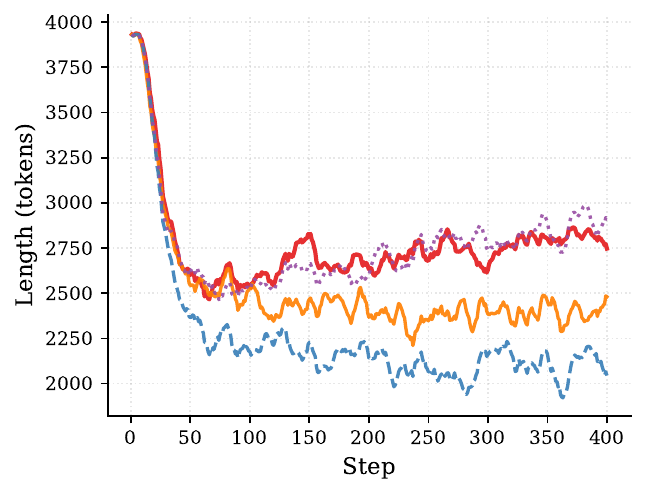}\\
        {\small (f) Response Length}
    \end{minipage}
    
    \caption{Ablation study on Qwen3-4B isolating the KL regularization ($\alpha$) and the full-vocabulary OPD term ($\beta$). The top row are the result of (a) AIME24, (b) AIME25, (c) AMC23 (all pass@1 mean@32). The bottom row shows the (d) reward, (e) entropy, (f) response length. URKL (default) and UFKL are the full SDPG variants, while ``$\alpha=0$'' removes the policy KL regularization and ``$\beta=0$'' removes the OPD term.}
    \label{fig:ablation_4b}
\end{figure}

The two ablations reveal complementary roles. Removing the full-vocabulary OPD term ($\beta=0$) preserves the reward and length profiles of URKL but loses the early-training accuracy advantage on AIME24 and AIME25 (Figures~\ref{fig:ablation_4b}a, b), confirming that privileged distillation is the primary driver of fast convergence on harder benchmarks. Removing the policy KL regularization ($\alpha=0$), in contrast, achieves comparable or slightly higher accuracy on AIME24/25 but at the cost of severely shortened response length (around 2{,}000 tokens, Figure~\ref{fig:ablation_4b}f) and rising entropy (Figure~\ref{fig:ablation_4b}e), suggesting that without the anchor the student begins to deviate from coherent reasoning patterns. These observations indicate that full-vocabulary OPD provides dense supervision, while the KL regularization stabilizes policy updates, so that we choose to retain both components.

\subsection{Robustness Across Model Scales: Qwen3-1.7B}
\label{appendix:ablation_1p7b}

\begin{figure}[ht!]
    \centering
    \includegraphics[width=0.75\linewidth]{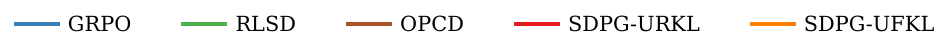}\\[4pt]
    
    \begin{minipage}[t]{0.32\linewidth}
        \centering
        \includegraphics[width=\linewidth]{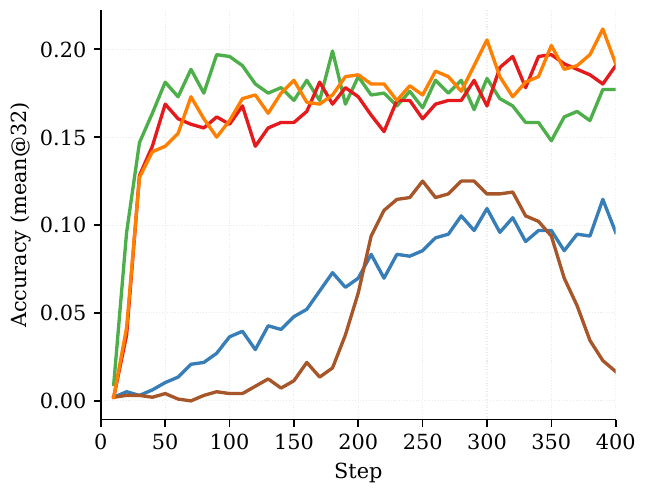}\\
        {\small (a) AIME24}
    \end{minipage}\hfill
    \begin{minipage}[t]{0.32\linewidth}
        \centering
        \includegraphics[width=\linewidth]{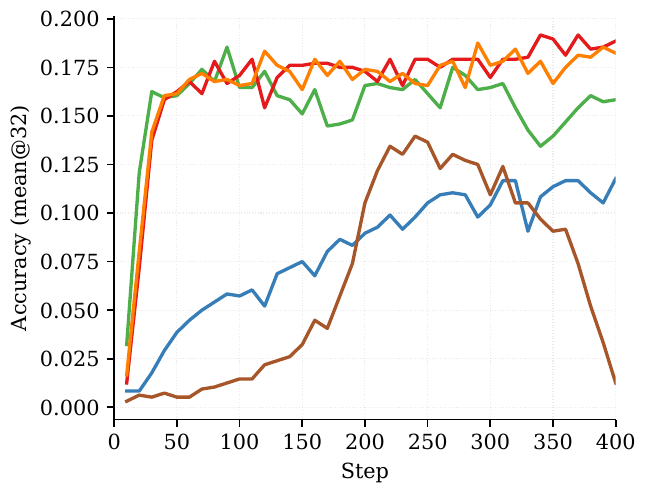}\\
        {\small (b) AIME25}
    \end{minipage}\hfill
    \begin{minipage}[t]{0.32\linewidth}
        \centering
        \includegraphics[width=\linewidth]{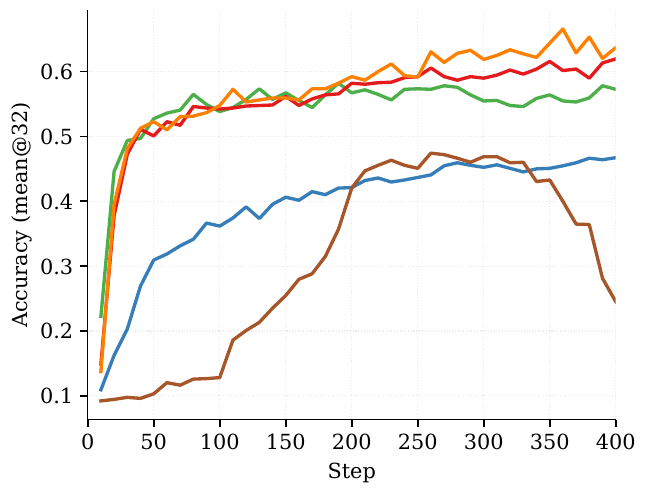}\\
        {\small (c) AMC23}
    \end{minipage}\\[8pt]
    
    \begin{minipage}[t]{0.32\linewidth}
        \centering
        \includegraphics[width=\linewidth]{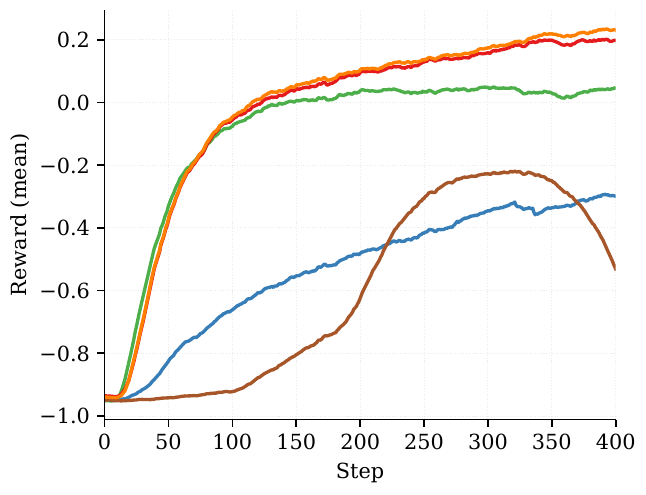}\\
        {\small (d) Reward}
    \end{minipage}\hfill
    \begin{minipage}[t]{0.32\linewidth}
        \centering
        \includegraphics[width=\linewidth]{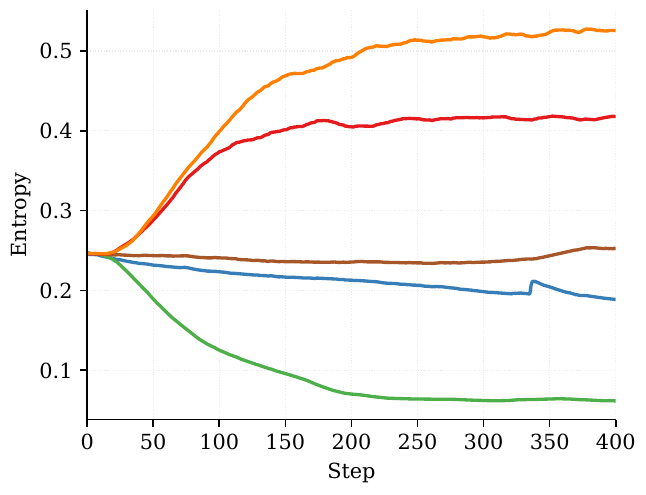}\\
        {\small (e) Entropy}
    \end{minipage}\hfill
    \begin{minipage}[t]{0.32\linewidth}
        \centering
        \includegraphics[width=\linewidth]{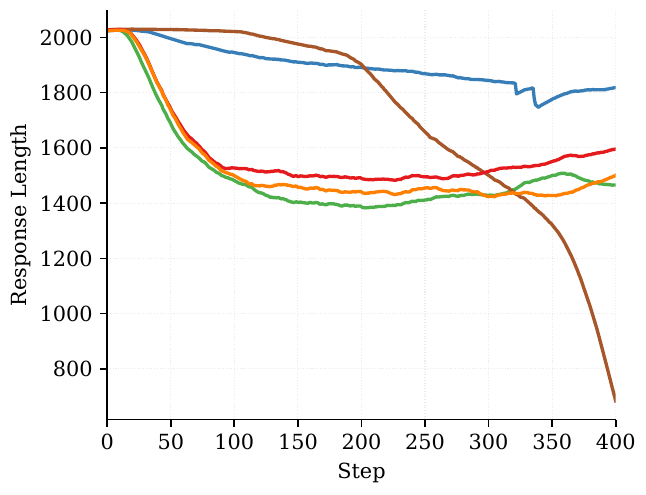}\\
        {\small (f) Response Length}
    \end{minipage}
    
    \caption{Training dynamics and benchmark performance on Qwen3-1.7B trained with baseline algorithms and SDPG variants. The top row are the results of: (a) AIME24, (b) AIME25, (c) AMC23 (all pass@1 mean@32). The bottom row are the results of (d) reward, (e) entropy, (f) response length. SDPG-URKL and SDPG-UFKL outperform GRPO and RLSD across all three benchmarks. OPCD, a pure self-distillation baseline without reward, exhibits training instability after step 250, with accuracy on AIME and response length both collapsing.}
    \label{fig:main_1p7b}
\end{figure}

To assess whether the design of SDPG generalizes beyond the 4B scale, we additionally run experiments on the smaller Qwen3-1.7B base model under the same training and evaluation protocol, with one additional baseline: OPCD~\citep{ye2026policy}. Results are shown in Figure~\ref{fig:main_1p7b} and Table~\ref{tab:main_1p7b}.

\begin{table}[ht!]
\centering
\caption{Performance on Qwen3-1.7B (pass@1, mean@32). \emph{Last} is the score at step 400; \emph{Best} is the peak across training. Column maximum in \textbf{bold}, second-best \underline{underlined}.}
\label{tab:main_1p7b}
\small
\begin{tabular}{l cc cc cc}
\toprule
& \multicolumn{2}{c}{\textbf{AIME24}} & \multicolumn{2}{c}{\textbf{AIME25}} & \multicolumn{2}{c}{\textbf{AMC23}} \\
\cmidrule(lr){2-3} \cmidrule(lr){4-5} \cmidrule(lr){6-7}
Method & Last & Best & Last & Best & Last & Best \\
\midrule
GRPO              & 0.096 & 0.115 & 0.118 & 0.118 & 0.467 & 0.467 \\
RLSD              & 0.177 & \underline{0.199} & 0.158 & 0.185 & 0.573 & 0.582 \\
OPCD              & 0.017 & 0.125 & 0.013 & 0.140 & 0.245 & 0.474 \\
\midrule
SDPG-URKL (ours) & \underline{0.191} & 0.197 & \textbf{0.189} & \textbf{0.192} & \underline{0.620} & \underline{0.620} \\
SDPG-UFKL (ours) & \textbf{0.192} & \textbf{0.212} & \underline{0.182} & \underline{0.188} & \textbf{0.637} & \textbf{0.666} \\
\bottomrule
\end{tabular}
\end{table}

The results on the 1.7B model corroborate the main findings at the 4B scale. SDPG-UFKL achieves the highest score on five of the six \emph{Last}/\emph{Best} columns, with SDPG-URKL taking the lead on AIME25, and the two SDPG variants together hold the top-two positions in every column except AIME24 \emph{Best}. The accuracy gap over GRPO and RLSD opens within the first 50 steps and persists throughout training (Figures~\ref{fig:main_1p7b}a--c). RLSD's entropy collapses below $0.1$ by step 200 (Figure~\ref{fig:main_1p7b}e), mirroring the pattern observed at the 4B scale, while SDPG-URKL and SDPG-UFKL maintain entropy above $0.4$ throughout training. Notably, OPCD, as a pure self-distillation baseline, exhibits sharp degradation after step 250, where the AIME24 accuracy drops from $0.13$ to $0.02$, response length collapses to under 300 tokens, and reward turns sharply negative (Figures~\ref{fig:main_1p7b}a, d, f). This instability supports the central design hypothesis of SDPG that full-vocabulary privileged distillation must be coupled with a binary outcome objective and a policy anchor to remain stable, especially on smaller models where pure imitation amplifies teacher imperfections.

\end{document}